\documentclass{article}
\usepackage{iclr2025_conference,times}

\usepackage{amsmath,amsfonts,bm}

\def\eqref#1{equation~\ref{#1}}

\def\1{\bm{1}}

\def\va{{\bm{a}}}
\def\vb{{\bm{b}}}
\def\vc{{\bm{c}}}

\def\vx{{\bm{x}}}
\def\vy{{\bm{y}}}

\DeclareMathAlphabet{\mathsfit}{\encodingdefault}{\sfdefault}{m}{sl}
\SetMathAlphabet{\mathsfit}{bold}{\encodingdefault}{\sfdefault}{bx}{n}

\def\gA{{\mathcal{A}}}
\def\gB{{\mathcal{B}}}
\def\gC{{\mathcal{C}}}
\def\gD{{\mathcal{D}}}

\def\gL{{\mathcal{L}}}

\def\gR{{\mathcal{R}}}
\def\gS{{\mathcal{S}}}

\def\gX{{\mathcal{X}}}
\def\gY{{\mathcal{Y}}}

\def\sP{{\mathbb{P}}}

\def\sV{{\mathbb{V}}}

\newcommand{\E}{\mathbb{E}}

\usepackage[T1]{fontenc}    
\usepackage{inconsolata}
\usepackage{hyperref}
\usepackage{url}
\usepackage{cancel}
\usepackage{graphicx}
\usepackage{booktabs}
\usepackage{wrapfig}
\usepackage{algorithm}
\usepackage{color}
\usepackage{bbm}
\usepackage{amsmath,amsfonts,amssymb,amsthm}
\usepackage[end]{algpseudocode}
\usepackage{appendix}
\usepackage[font=small,labelfont=bf]{caption}
\usepackage[thinc]{esdiff}  
\usepackage{pifont}
\usepackage{bbding}
\usepackage{tabularray}
\usepackage{titlesec}
\usepackage{scalerel,stackengine}
\usepackage{mathtools}
\usepackage{transparent}
\usepackage{tikz}
\usetikzlibrary{fit,calc}
\usepackage{enumitem}
\usepackage{multirow}
\usepackage{tcolorbox}
\usepackage{colortbl}
\usepackage{xcolor}
\usepackage{fontawesome5}

\titlespacing*{\section}
{0pt}{1ex}{1ex}
\titlespacing*{\subsection}
{0pt}{1ex}{1ex}

\definecolor{mydarkblue}{rgb}{0,0.08,0.45}
\definecolor{mydarkgreen}{RGB}{0, 139, 69}
\definecolor{MAEblue}{RGB}{47 112 182}
\definecolor{SDEblue}{RGB}{28 58 88}
\definecolor{Refcolor}{RGB}{47 112 182}
\hypersetup{
 colorlinks=true,
 urlcolor=magenta,
 citecolor=SDEblue,
}
\definecolor{mycyan}{cmyk}{.3,0,0,0}

\newcommand{\cmark}{\ding{51}}%
\newcommand{\aref}[1]{\hyperref[#1]{Appendix~\ref*{#1}}}

\hypersetup{linkcolor = Refcolor}

\usepackage{cleveref} 
\crefname{section}{Section}{Sections}
\crefname{theorem}{Theorem}{Theorems}
\crefname{corollary}{Corollary}{Corollaries}
\crefname{lemma}{Lemma}{Lemmas}
\crefname{equation}{Eq.}{Eq.}
\crefname{proposition}{Proposition}{Propositions}
\crefname{claim}{Claim}{Claims}
\crefname{remark}{Remark}{Remarks}
\crefname{observation}{Observation}{Observations}
\crefname{assumption}{Assumption}{Assumptions}
\crefname{definition}{Definition}{Definitions}
\crefname{appendix}{Appendix}{Appendices}
\crefname{algorithm}{Algorithm}{Algorithms}
\crefname{figure}{Figure}{Figures}
\crefname{table}{Table}{Tables}
\crefname{property}{Property}{Properties}
\crefname{line}{Line}{Lines}

\newtheorem{property}{Property}

\definecolor{logocolor}{RGB}{30, 0, 178}      
\newcommand{\sea}{{\textcolor{logocolor}{\textbf{\textsc{\small{SEA}}}}}}
\newcommand{\seatitle}{{\textcolor{logocolor}{\textbf{\textsc{{SEA}}}}}}
\newcommand{\piref}{\pi_\mathrm{ref}}
\newcommand{\highlight}[1]{{\color{red!20!violet}#1}}

\title{Sample-Efficient Alignment for LLMs}

\newcommand{\sail}{1}
\newcommand{\nus}{2}
\newcommand{\smu}{3}
\newcommand{\sailnus}{1,2}
\newcommand{\sailsmu}{1,3}

\renewcommand\footnotemark{}
\author{
Zichen Liu\textsuperscript{\sailnus} \quad
Changyu Chen\textsuperscript{\sailsmu} \quad
Chao Du\textsuperscript{\sail$\dagger$}\thanks{$^\dagger$Corresponding author.} \quad
Wee Sun Lee\textsuperscript{\nus}\quad 
Min Lin\textsuperscript{\sail} \quad \\
\textsuperscript{\sail}Sea AI Lab \quad
\textsuperscript{\nus}National University of Singapore \quad
\textsuperscript{\smu}Singapore Management University
\\ \texttt{\{liuzc,chency,duchao,linmin\}@sea.com} \quad
\texttt{\{zichen,leews\}@comp.nus.edu.sg}
}

\iclrfinalcopy 

\begin{document}
\maketitle

\begin{abstract}
We study methods for efficiently aligning large language models (LLMs) with human preferences given budgeted online feedback. We first formulate the LLM alignment problem in the frame of contextual dueling bandits. This formulation, subsuming recent paradigms such as online RLHF and online DPO, inherently quests for sample-efficient algorithms that incorporate \textit{online active exploration}. Leveraging insights from bandit theory, we introduce a unified algorithm based on {\textbf{Thompson sampling}} and highlight its applications in two distinct LLM alignment scenarios. The practical agent that efficiently implements this algorithm, named \sea~(\textbf{S}ample-\textbf{E}fficient \textbf{A}lignment), is empirically validated through extensive experiments across three model scales (1B, 2.8B, 6.9B) and three preference learning algorithms (DPO, IPO, SLiC). The results demonstrate that \sea~achieves highly sample-efficient alignment with oracle's preferences, outperforming recent active exploration methods for LLMs. Additionally, we release the implementation of \sea~together with an efficient codebase designed for \textbf{\underline{o}}nline \textbf{\underline{a}}lignmen\textbf{\underline{t}} of LLMs, aiming to accelerate future research in this field.

\vspace{0.3cm}
\centering{
\faGithub~\url{https://github.com/sail-sg/oat}
}
\end{abstract}

\begin{figure}[!htb]
    \vspace{-0.2em}
    \begin{minipage}{0.49\textwidth}
        \centering
        \includegraphics[width=0.9\linewidth]{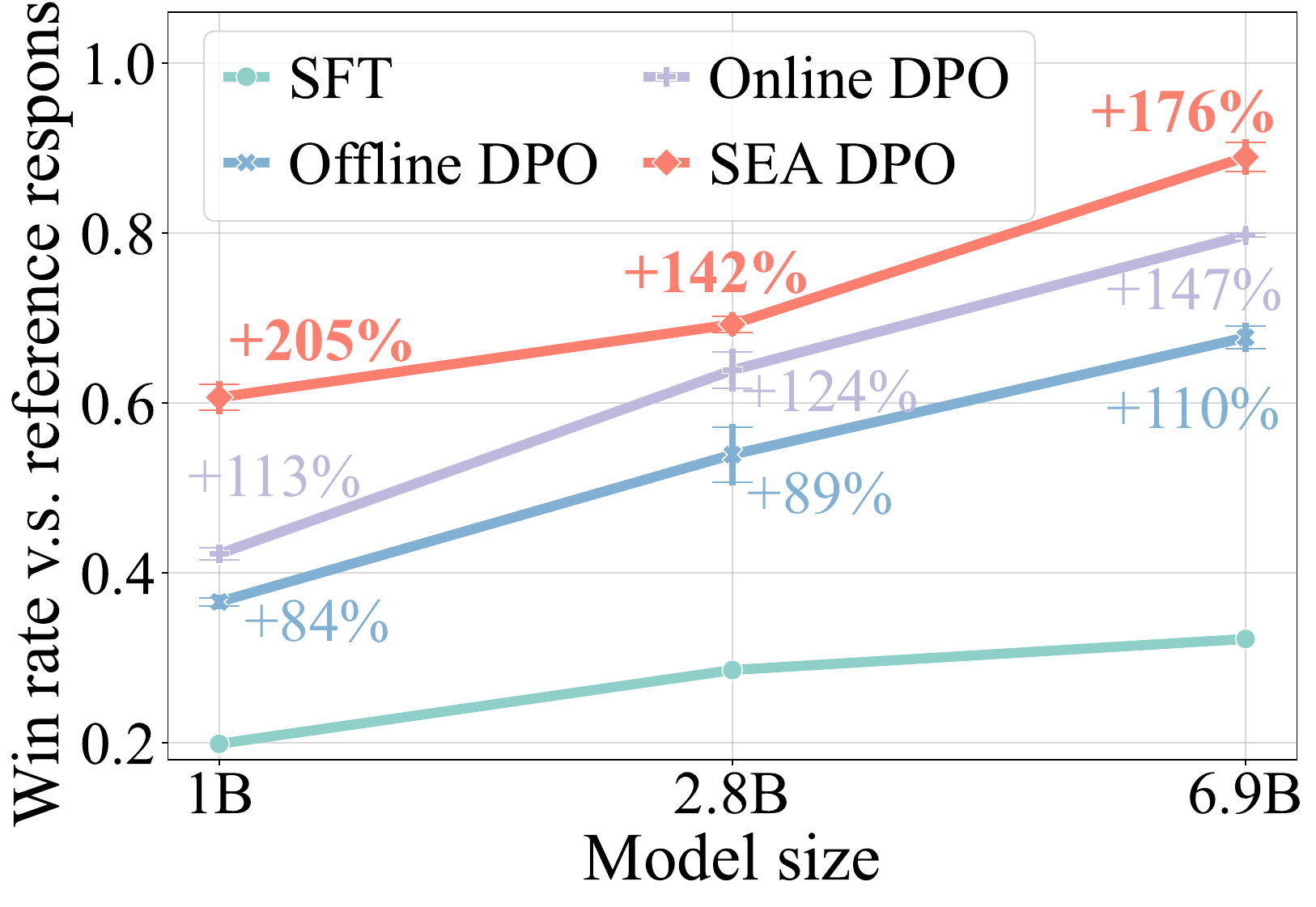}
    \end{minipage}
    \begin{minipage}{0.49\textwidth}
        \centering
        \includegraphics[width=0.9\linewidth]{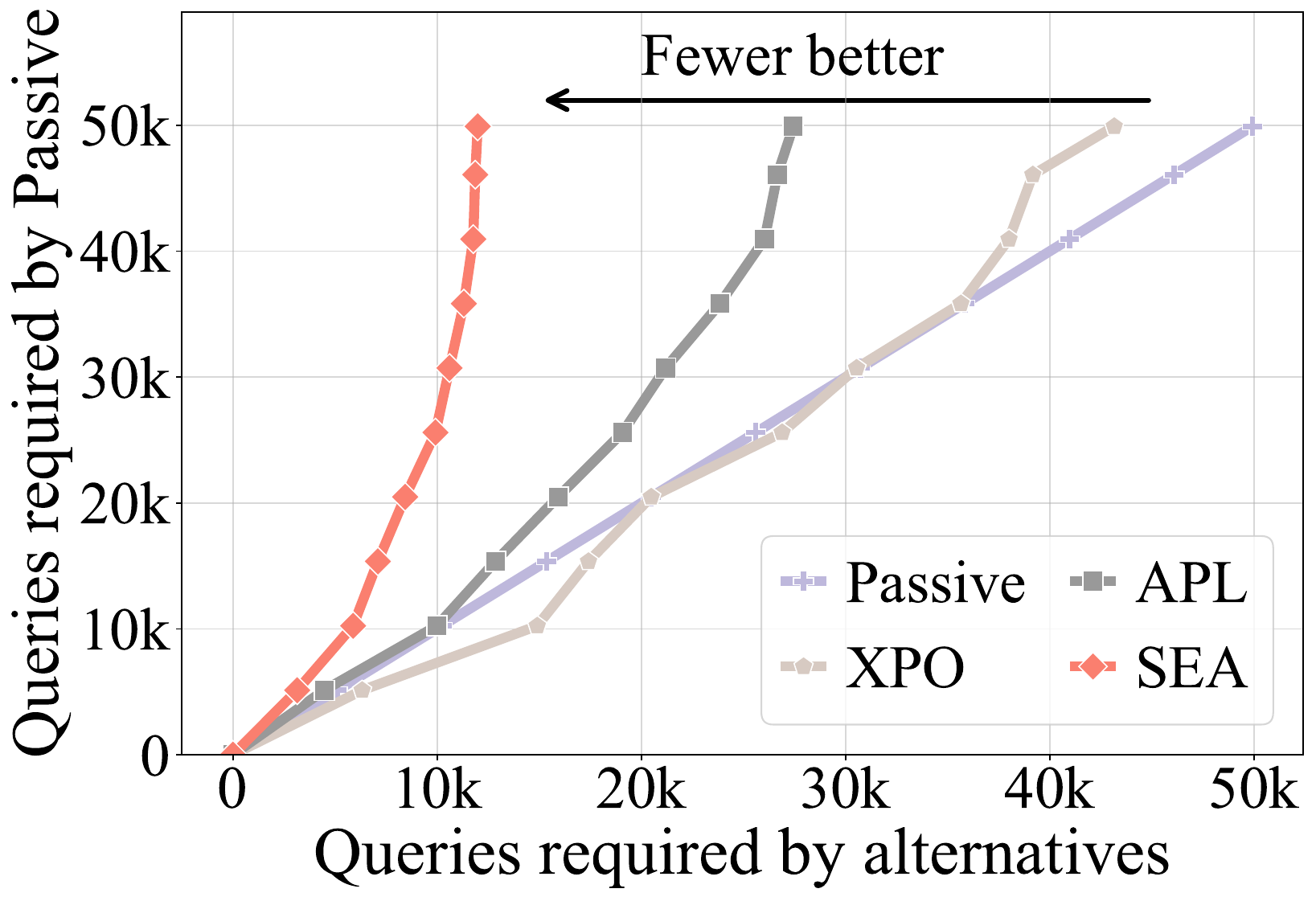}
    \end{minipage}
    \caption{Win rate comparison of model responses against reference responses on the {TL;DR} task, judged by the preference oracle. All compared methods use the same optimization method (DPO).
    \textbf{(Left)} Performance improvements at convergence over {SFT} models achieved by offline ({Offline DPO}), passively online ({Online DPO}), and our \textit{active exploration} (\sea{} {DPO}) methods. \textbf{(Right)} The number of queries required by the passively online method ({Passive}) versus that by different active exploration methods to attain various levels of win rates. \sea~achieves the best sample efficiency for online alignment compared to {XPO} and {APL}.}
    \label{fig:3-scale-benchmark}
\end{figure}

\section{Introduction}

Aligning LLMs with human preferences is a crucial step to elicit various desirable behaviors, e.g., helpfulness and harmlessness~\citep{bai2022training}. Moreover, it holds the potential to create superhuman capabilities with only human-level feedback, as verifying is believed to be easier than synthesizing novel behaviors. By iteratively generating massive new candidates and asking for human feedback, LLMs could learn to reinforce good behaviors and may eventually surpass human capabilities.\looseness=-1

Existing methods, either via reinforcement learning from human feedback (RLHF)~\citep{stiennon2020learning,ouyang2022training} or direct alignment from preferences (DAP)~\citep{rafailov2023dpo,azar2024ipo}, typically require a large amount of human annotations to achieve effective alignment. As a result, the volume of human feedback becomes a major bottleneck in practical alignment scenarios. This poses a challenging and under-explored research question:
\begin{center}
    \textit{\textbf{How to align LLMs sample-efficiently?}}
\end{center}

To seek answers, in \cref{sec:formulation}, we formalize LLM alignment as a contextual dueling bandit (CDB)~\citep{yue2012k,dudik2015contextual}, where the {agent} (i.e., the learner and decision maker, in our case the LLM) interacts with the environment (i.e., human) to collect experience for improving its policy.
This formulation naturally calls for two key properties for alignment algorithms to be sample-efficient:
\begin{property}[{online interaction}]
    \label{property1}
    \textnormal{Interacting and learning \textit{online} allows the agent to act with the latest learned policy and then use that experience to immediately improve the policy.}
\end{property}
\begin{property}[{active exploration}]
    \label{property2}
    \textnormal{An \textit{actively exploring} agent strategically selects actions such that the collected experience leads to maximal policy improvement.}
\end{property}
Since the CDB formulation is general and almost subsumes all existing LLM alignment methods, it provides us a lens to scrutinize prior methods on the axes of \cref{property1,property2}. In \cref{sec:existing_work}, we thoroughly discuss prior alignment approaches, ranging from offline learning~\citep{rafailov2023dpo,azar2024ipo} and passive learning with iterative~\citep{christiano2017deep,dong2024rlhf} or online interaction~\citep{guo2024direct}, to active exploration for learning preference models~\citep{dwaracherla2024efficient} or aligning LLMs~\citep{muldrew2024active,zhang2024self,xie2024exploratory}. As will be revealed, most prior methods (partially) fail to satisfy the two properties, resulting in inferior sample efficiency. Moreover, through the CDB formulation, we identify two LLM alignment scenarios, namely \textcolor{black}{aligning from online users' feedback} (e.g., \citet{chatgpt}) and aligning from crowdsourcing~\citep{christiano2017deep,ouyang2022training}, and shed light on their correspondences to two bandit settings (\textcolor{black}{explore \& exploit} and best arm identification). Understanding their differences is important for designing efficient alignment algorithms for respective scenarios. We detail these two settings in \cref{sec:formulation} and discuss how prior works approach them in \cref{sec:existing_work}.\looseness=-1

Leveraging algorithmic insights from bandit theory, our answer to the research question above is a principled alignment algorithm based on Thompson sampling (TS)~\citep{thompson1933likelihood}. Our method fulfills \cref{property1,property2} to enhance sample efficiency, and it solves either of the two settings depending on practical scenarios (\cref{sec:ts_llm_alignment}). We incorporate techniques including \textit{epistemic reward model}, \textit{policy-guided search} and \textit{mixed preference learning} to implement the proposed TS algorithm (\cref{sec:sea_impl}), yielding a practical agent which we call \sea~(\textbf{S}ample-\textbf{E}fficient \textbf{A}lignment). In addition, we develop and open source a highly efficient, distributed learning system for studying online LLM alignment methods (\cref{sec:setup}), eliminating barriers to \textit{fair} empirical comparisons of different alignment algorithms. Through extensive experiments (\cref{sec:empirical_results}), \sea~shows strong empirical results (see \cref{fig:3-scale-benchmark}), consistently achieving higher win rates and improved sample efficiency compared to baseline approaches across three model scales. We hope our open-sourced codebase and proposed algorithm could inspire future research in sample-efficient online LLM alignment.

\vspace{-0.1cm}
\section{LLM alignment as contextual dueling bandits}
\label{sec:formulation}
\vspace{-0.1cm}
We first review the definitions and two typical objectives of \textit{Contextual Dueling Bandits} (\cref{sec:formulation:cdb}), then translate them into the language of \textit{LLM alignment} (\cref{sec:formulation:alignment}). The tight connection between them, as we will see, allows us to leverage insights from bandit algorithms to design efficient alignment algorithms for LLMs.

\vspace{-0.1cm}
\subsection{Contextual dueling bandits}
\label{sec:formulation:cdb}
\vspace{-0.1cm}

Contextual dueling bandits (CDB)~\citep{yue2012k, dudik2015contextual} is proposed to study online learning problems where the feedback consists of relative pairwise comparisons. A CDB problem can be characterized by a tuple $(\gC, \gA, \sP)$, where $\gC$ is the context space, $\gA$ is the action space, and $\sP: \gA \times \gA \times \gC \mapsto [0, 1]$ denotes the unknown \textit{preference oracle}. An agent learns by iteratively interacting with the environment (i.e., the preference oracle $\sP$) as follows. At each round $t$ of the learning process, a context $\vc_t\sim p_\gC$ is presented to the agent, who needs to take two actions $\va_t, \va_t' \in \mathcal{A}$ for a ``dueling'' comparison. The agent then receives stochastic feedback in the form of a comparison result $z_t \sim \operatorname{Ber}\left(\mathbb{P}\left(\va_t \succ \va_t'| \vc_t \right) \right)$ from the environment, where $\operatorname{Ber}(\cdot)$ is the Bernoulli distribution and $\succ$ denotes that the first action is preferred. 

\textbf{Regret}. The quality of the dueling actions selected by the agent is measured by the \textit{immediate regret}: $R_t = \sP(\va^\star_{t} \succ \va_t | \vc_t) + \sP(\va^\star_{t} \succ \va_t' | \vc_t) - 1$, where $\va^\star_t$ is the best action\footnote{We assume that a best action $\va^\star$ in the sense that $\sP(\va^\star \succ \va | \vc) \geq \frac{1}{2},\forall \va\in\gA$ exists for all context $\vc\in\gC$.} the agent would take at round $t$ if it had complete knowledge of $\sP$.
Intuitively, if the agent has learned how to act optimally from round $t$ onwards, it would no longer suffer any regret since its actions would be indistinguishable from the best action ($\sP(\va_\tau^\star \succ \va_\tau | \vc_\tau) = \frac{1}{2}$ hence $R_\tau=0$ for $\tau \geq t$).

\textbf{Optimal policy}.
A policy $\pi \in \Delta^\gC_\gA$\footnote{We denote by $\Delta^\gC_\gA$ the set of all mappings $\gC\mapsto\Delta_\gA$, where $\Delta_\gA$ denotes the set of all probability distributions over $\gA$.} associates each context $\vc\in \gC$ with a probability distribution $\pi(\cdot|\vc) \in \Delta_\gA$ over the action space.
The \textit{total preference} of policy $\pi$ over policy $\mu$ given a context sampling distribution $p_\gC\in\Delta_\gC$ and a preference oracle $\sP$ is defined as
\begin{equation}
    \label{eq:total_preference}
    P_{p_\gC, \sP}(\pi \succ \mu) = {\mathbb{E}}_{\vc \sim p_\gC}\left[{\mathbb{E}}_{\va \sim \pi(\cdot | \vc)}{\mathbb{E}}_{\va' \sim \mu(\cdot | \vc)}\left[\sP(\va \succ \va' | \vc)\right]\right].
\end{equation}
We adopt the \textit{von Neumann winner}~\citep{dudik2015contextual} as the solution concept, which requires the optimal policy $\pi^\star$ to satisfy that
\begin{equation}
    \label{eq:cdb_optimal_pi}
    \forall \pi' \in \Delta^\gC_\gA, \; P_{p_\gC, \sP}(\pi^\star \succ \pi') \geq \frac{1}{2}.
\end{equation}
In words, the von Neumann winner policy should beat or tie with every policy (i.e., is zero-regret) on average.

\textbf{Learning objectives.} The goal of bandit agents is to learn an optimal policy through interactions with the environment. There are two subtypes of objectives that focus on different learning scenarios.
The first type considers the conventional \textit{explore and exploit (E\&E)} setting~\citep{robbins1952sequential, auer2002finite}, where the agent learns fully \textbf{online} and tries to minimize the cumulative regret over $T$ rounds: $\sum_{t=1}^T R_t$. The second type of objective concerns the \textit{best arm identification (BAI)} setting~\citep{bubeck2009pure,audibert2010best}, where the agent is only evaluated \textbf{offline} on its average performance, possibly at any round (a.k.a., anytime regret), and tries to learn the optimal policy with minimum interaction. Both settings call for effective \textit{online exploration} strategies that satisfy \cref{property1,property2}. Their differences will be made clearer with real scenarios in \cref{sec:formulation:alignment}.\looseness=-1

\vspace{-0.1cm}
\subsection{Alignment as CDB}
\label{sec:formulation:alignment}
\vspace{-0.1cm}

\begin{figure}[t]
    \centering
    \includegraphics[width=0.9\linewidth]{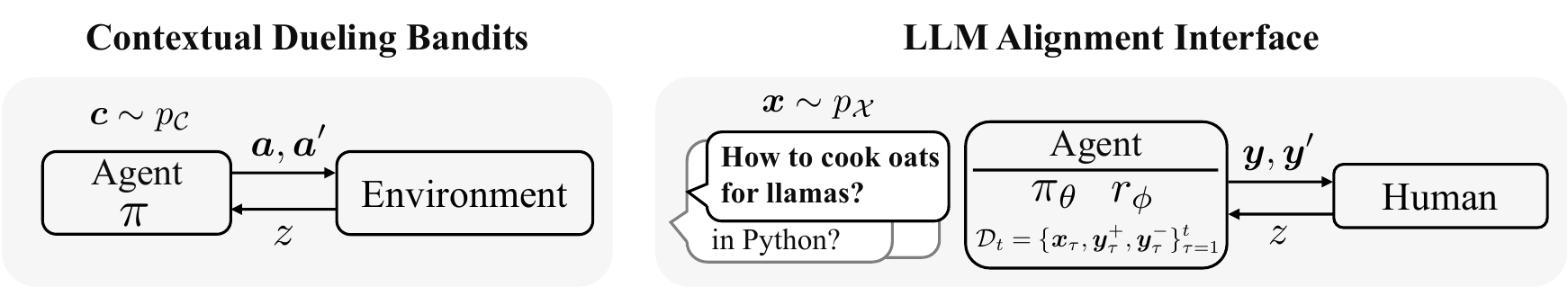}
    \caption{Illustrative comparison between CDB and LLM alignment.}
    \label{fig:interface}
\end{figure}
LLM alignment can be framed as a CDB problem with their correspondences illustrated in \cref{fig:interface}. Specifically, at time $t$ a text prompt (cf.\ context) $\vx_t \in \gX$ is sampled from a prompt distribution $p_\mathcal{X}$. Then, two distinct responses (cf.\ actions), $\vy_t, \vy_t' \in \gY$, are chosen by the agent, and presented to human annotators (cf.\ the environment) for preference ranking. The winning and losing responses are labeled as $(\vy_t^+, \vy_t^-)$ based on a binary stochastic feedback $z_t$. The agent is expected to behave optimally by pursuing either E\&E or BAI objectives, with knowledge learned from the experience accumulated so far: $\mathcal{D}_t = \{\bm{x}_\tau, \bm{y}_\tau^+, \bm{y}_\tau^-\}_{\tau=1}^t$. A standard assumption is that human preferences follow the Bradley-Terry (BT) model \citep{bradleyterry1952rank}:
\begin{equation}
    \label{eq:bt_model}
    \mathbb{P}(\vy_t \succ \vy_t'|\vx_t) = \frac{\exp{(r^\star(\vx_t, \vy_t))}}{\exp{(r^\star(\vx_t, \vy_t))} + \exp{(r^\star(\vx_t, \vy_t'))}} = \sigma(r^\star(\vx_t, \vy_t) - r^\star(\vx_t, \vy_t')),
\end{equation}
where $\sigma$ is the sigmoid function and $r^\star$ encodes human's implicit reward. The immediate regret of LLM alignment can be rewritten as $R_t = r^\star(\vx_t, \vy_t^\star) - \left({r^\star(\vx_t, \vy_t) + r^\star(\vx_t, \vy_t')}\right)/{2}$ with the BT assumption~\citep{saha2021optimal,li2024fgtscdb}, where $\vy_t^\star$ is the best response for prompt $\vx_t$ given human's implicit reward, i.e., $r^\star(\vx_t,\vy_t^\star) \geq r^\star(\vx_t,\vy),\forall \vy\in\gY$. The von Neumann winner policy is also redefined as
\begin{equation}
    \label{eq:optimal_policy}
    \pi^\star \in \underset{\pi\in \Delta^\gX_\gY}{\arg\max}~J(\pi), \text{ where } J(\pi)={\E}_{\vx \sim p_\gX}{\E}_{\vy \sim \pi(\cdot | \vx)}{\left[ r^\star(\vx, \vy) \right]} ~\text{is the objective,}
\end{equation}
by substituting \cref{eq:bt_model} into \cref{eq:total_preference} and maximizing $P_{p_\gX, \sP}(\pi \succ \pi^\star)$ towards $1/2$.

\label{para:different_obj}\label{sec:llm_objectives} The \textbf{two settings in bandits} have their respective applications in LLM alignment. \textbf{\highlight{(1)}} The E\&E setting applies to the scenario of serving an LLM-based application online and aligning it continually with users' preferences. In this setting, the agent needs to balance exploration with exploitation, thus the cumulative regret is of interest because the quality of \textit{every} response matters. In fact, commercial systems like ChatGPT would strategically ask users to make a dueling comparison, while upholding the quality of both responses. Please see \cref{fig:chat-gpt-online-feedback} in \cref{sec:app:supp} for an example. \textbf{\highlight{(2)}} The BAI setting corresponds to the other scenario where annotators are paid to provide human feedback~\citep{christiano2017deep, ouyang2022training}. The desideratum in this scenario is to align the LLM at the minimum labeling cost, while the quality of the dueling responses is not important as long as the experience helps sample-efficiently learn the von Neumann winner policy.

After formalizing LLM alignment in the framework of CDB and uncovering their tight connections, we next thoroughly discuss existing alignment methods in the CDB framework and reveal the sources of their sample inefficiencies.

\vspace{-0.1cm}
\section{How prior works (partially) solve LLM alignment as CDB}
\vspace{-0.1cm}
\label{sec:existing_work}

\definecolor{cmpblue}{RGB}{0,120,175}
\definecolor{cmpgreen}{RGB}{45,131,15}
\definecolor{cmppurple}{RGB}{125,12,113}
\definecolor{cmporange}{RGB}{186,70,9}
\begin{figure}
    \centering
    \includegraphics[width=\linewidth]{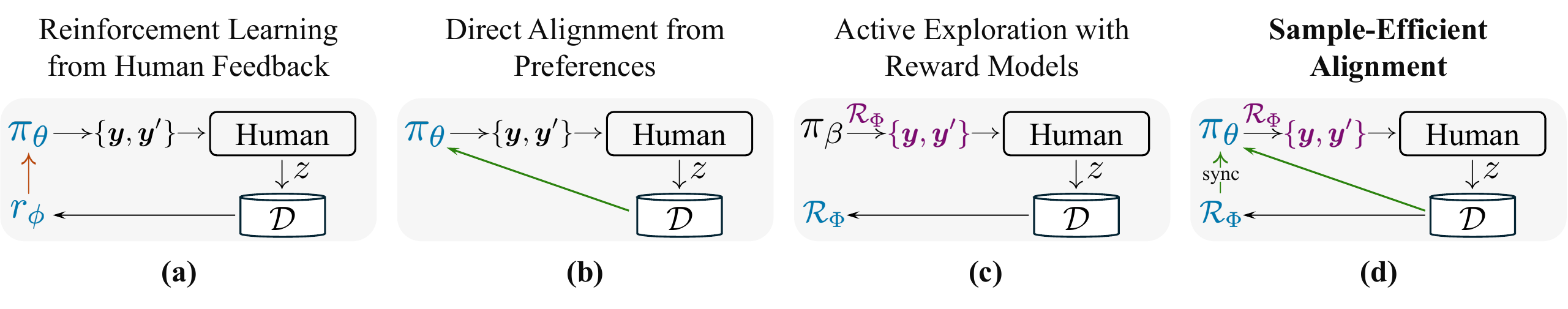}
    \caption{Different paradigms for solving the LLM alignment problem in the CDB framework. Note that although all paradigms follow the LLM alignment interface (\cref{fig:interface}) with the interaction loop, some are actually offline or iteratively online (i.e., loop only once or a few times). Detailed comparisons will be made in \cref{sec:existing_work}. We use colors to denote \textcolor{cmpblue}{learnable components}, \textcolor{cmporange}{RL optimizer}, \textcolor{cmpgreen}{direct optimizer}, and \textcolor{cmppurple}{active exploration}. $r_\phi$ denotes a point estimate of human's implicit reward, while $\gR_\Phi$ refers to an uncertainty-aware reward model.\looseness=-1}
    \label{fig:compare}
\end{figure}

We first align the notations and terminology used in CDB with commonly referred ones in the LLM community. Previously, we used the term ``agent'' to denote the learner and decision maker, and referred to its overall behavior as the ``policy'' $\pi$ (as in \cref{eq:optimal_policy}), following the standard abstraction in RL~\citep{sutton2018rlbook,sutton2022alberta}.
However, in the LLM literature, ``policy'' typically refers to the generative language model alone, excluding components like reward models (RMs) that the agent might additionally build (see \cref{fig:interface}). To avoid confusion, from now on we use $\pi_{\theta^t}$ to denote the generative language model (policy) and $r_{\phi^t}$ to denote the (optional) RM at time $t$, both of which are learned from preference data $\gD_t$ collected up to time $t$. We will omit $t$ when the time-indexing is not applicable (i.e., no online interaction) or not important in the context.

\textbf{RLHF and DAP}. Commonly adopted RLHF pipelines~\citep{christiano2017deep,stiennon2020learning,bai2022training,ouyang2022training} first learn a proxy RM with a negative log-likelihood loss:\looseness=-1
\begin{equation}
\label{eq:r_loss}
    \gL_r(\phi|\mathcal{D}) = - \E_{\left(\vx, \vy^+, \vy^-\right) \sim p_\mathcal{D}}\left[\log \sigma\left(r_\phi\left(\vx, \vy^+\right)-r_\phi\left(\vx, \vy^-\right)\right)\right],
\end{equation}
where $\gD$ is collected by querying human annotators using a behavior policy $\piref$ (typically the supervised fine-tuned policy $\pi_{\mathrm{sft}}$). Afterwards, \textit{offline RL}~\citep{lange2012batch,levine2020offline} is conducted to learn $\pi_\theta$ with respect to the learned reward $r_{\phi}$ \textcolor{cmporange}{internally} within the agent~(\cref{fig:compare}{\color{Refcolor}a}). However, the learned model $\pi_{\theta}$ might be inaccurate at regions out of the distribution (o.o.d.) of $\piref$ because little training data can be collected. An effective remedy is to incorporate a pessimistic term to combat the distributional shift, leading to a reformulation of the von Neumann winner policy objective in \cref{eq:optimal_policy} as
\begin{align}
    J(\pi_\theta)&=\underset{\vx \sim p_\gX}{\E}~\underset{\vy \sim \pi_\theta(\cdot | \vx)}{\E}{\Biggr[ \underbrace{\biggr.r_\phi(\vx, \vy)}_{\text{~~estimated $r^\star$}} - \underbrace{\beta\log\frac{\pi_\theta(\vy|\vx)}{\piref(\vy|\vx)}\biggl.}_{\text{o.o.d.\ reward penalty~}}~\Biggl]} \\
    \label{eq:kl_reg_obj}&= \underset{\vx \sim p_\gX}{\E}\left[ \underset{\vy \sim \pi_\theta(\cdot | \vx)}{\E} \left[ r_\phi(\vx, \vy) \right] - \beta D_{\text{KL}}(\pi_\theta(\cdot|\vx) || \piref(\cdot|\vx)) \right],
\end{align}
which converts an online objective regarding the human's implicit reward $r^\star$ to an offline objective regarding the proxy reward $r_\phi$. The KL penalty in \cref{eq:kl_reg_obj} is widely used for language model fine-tuning~\citep{jaques2020human,xiong2024iterative}, and PPO~\citep{schulman2017proximal} has become a default \textcolor{cmporange}{\textit{RL optimizer}} to maximize the KL-regularized reward. However, the performance of RLHF is guaranteed only if the preference data $\gD$ induced by $\piref$ adequately covers $\pi^\star$~\citep{zhu2023principled}, which is often approximated by updating $\piref$ with the latest (improved) $\pi_\theta$ for re-sampling a batch of online experience and repeating \cref{eq:r_loss,eq:kl_reg_obj}. Prior methods typically employ only a few iterations of online interaction with large batches~\citep{xiong2024iterative,dong2024rlhf}, which may compromise sample efficiency (\cref{property1}).

Online RLHF is difficult due to the complexity and instability of RL optimizers. For example, \citet{huang2024n+} openly reproduces offline RLHF scaling behaviors but requires many implementation tricks for training, highlighting the difficulties of an online counterpart. Fortunately, the introduction of DAP (or \textcolor{cmpgreen}{\textit{direct optimizers}}) largely simplifies and stabilizes the alignment process by conducting contrastive supervised learning directly on $\gD$ (\cref{fig:compare}{\color{Refcolor}b}). While most DAP works focus on learning from a fixed offline preference dataset~\citep{zhao2023slic,rafailov2023dpo,azar2024ipo,meng2024simpo,zhang2024chain}, iterative DPO~\citep{xu2023iterativedpo} observes improved results when allowing iterative online interaction.~\citet{guo2024direct} further propose OAIF to make DAP faithfully online, satisfying \cref{property1}, and demonstrate that online learning prevents over-fitting and yields continual performance improvement. Nevertheless, it still employs passive exploration strategies (directly using $\vy,\vy' \sim \pi_\theta$), hindering sample efficiency (\cref{property2}).

\textbf{Online exploration in LLMs}. A line of recent works~\citep{mehta2023sample,das2024provably, melo2024deep,dwaracherla2024efficient} adopts the fully online bandit formulation and incorporates \textcolor{cmppurple}{\textit{active exploration}} with uncertainty-aware RMs for response selection~(\cref{fig:compare}{\color{Refcolor}c}). In particular,~\citet{mehta2023sample} consider the E\&E setting and develop a UCB-style~\citep{auer2002finite} algorithm; \citet{das2024provably} instead select the dueling responses with the most uncertain preference estimate, targeting the BAI setting in a pure exploration way; unlike the above, \citet{melo2024deep} view the problem from the angle of pool-based active learning and propose an acquisition function based on both entropy and epistemic uncertainty; finally, the work by \citet{dwaracherla2024efficient} is the closest to ours in the sense that they apply double Thompson sampling (DTS)~\citep{wu2016double} for exploration, but DTS is designed for the E\&E setting while they evaluate anytime average performance as in the BAI setting. We will show in \cref{sec:choice_of_exploration} that pure exploration by~\citet{das2024provably} is not the best choice for BAI, and the objective mismatch in~\citet{dwaracherla2024efficient} could lead to suboptimal performance in respective settings. Meanwhile, all these works primarily focus on learning uncertainty-aware RMs online without updating LLM policies. Therefore, all responses are sampled from a fixed proposal policy $\pi_\beta$ (or even a fixed dataset), making the data coverage a critical concern.

Another line of research updates LLMs online while incorporating exploration. \citet{zhang2024self} and \citet{xie2024exploratory} independently propose to learn an optimistic RM to encourage exploration. They leverage the property of DPO~\citep{rafailov2023dpo} to reparameterize RM with policy and conclude with an extra optimistic term in the DPO loss function. Thus, their learning processes are like \cref{fig:compare}{\color{Refcolor}b} but with an optimistic direct optimizer. \citet{muldrew2024active} adopt the vanilla DPO loss but utilize the implicit reward margin to actively select dueling responses. Yet, these methods are tightly coupled with DPO and not compatible to other direct optimizers. Their experiments are also limited to a few online iterations, possibly due to the implementation difficulty of a faithfully online learning system. Given their relevance to our approach, we will reproduce them in a fully online manner for fair comparisons in \cref{sec:overall_cmp}. We summarize prior works in \cref{tab.compare} in \cref{sec:app:supp}.

\section{\seatitle: sample-efficient alignment for LLMs}
\label{sec:sea}

In this section we present our online exploration agent~\sea~(\cref{fig:compare}{\color{Refcolor}d}). We first introduce a principled Thompson sampling algorithm inspired by bandit theory (\cref{sec:ts_llm_alignment}), and then derive~\sea~as its practically efficient implementation (\cref{sec:sea_impl}). Interestingly,~\sea~can also be viewed as an instantiation of a classical model-based RL architecture called Dyna~\citep{sutton1990dyna}, for which we defer the discussion to \cref{sec:app:dyna}.

\vspace{-0.1cm}
\subsection{Thompson sampling for LLM alignment}
\label{sec:ts_llm_alignment}
\vspace{-0.1cm}
Thompson sampling (TS)~\citep{thompson1933likelihood} is widely adopted for solving bandit problems at scale due to its efficiency and strong empirical performance in general online learning problems~\citep{chapelle2011empirical,russo2018tutorial}. A bandit agent using Thompson sampling typically maintains and incrementally updates a posterior distribution of the oracle reward $p(r|\gD)$. Meanwhile, the agent takes actions following a greedy policy with respect to a sampled RM: $\va_t = \arg\max_{\va} r(\va)$ with $r \sim p_r(\cdot|\gD)$. This simple yet effective algorithm naturally balances exploration and exploitation: when the agent has limited knowledge about the environment, the posterior estimate exhibits high uncertainty so that the sampled RM could guide the greedy policy to explore; after sufficient experience is gathered, the sampled RM approximates the oracle more closely, allowing the agent to exploit near-optimal policies.

In the context of LLM alignment, we leverage the BT assumption (\cref{eq:bt_model}) to replace the preference oracle $\sP$ with the implicit reward $r^\star$. This substitution enables us to model the reward posterior $p(r|\gD)$ in the standard TS framework, preserving the probabilistic structure necessary for effective posterior sampling. Inspired by prior works~\citep{wu2016double,gonzalez2017preferential} on non-contextual $K$-arm bandits and preferential Bayesian optimization problems, we generalize them for LLM alignment and develop a unified algorithm as shown in \cref{algo:theoretical_algo}. Note that we assume for now the LLM agent can be fully described by the posterior $p(r|\gD)$, and we defer practical reward ($r_\phi$) and policy ($\pi_\theta$) learning to \cref{sec:sea_impl}.

\newcommand*{\tikzmk}[1]{\tikz[remember picture,overlay,] \node (#1) {};\ignorespaces}
\newcommand{\boxitee}[1]{\tikz[remember picture,overlay]{\node[yshift=2pt,xshift=-7pt,fill=#1,opacity=.15,fit={(A)($(B)+(.65\linewidth,.8\baselineskip)$)}] {};}\ignorespaces}
\newcommand{\boxitbai}[1]{\tikz[remember picture,overlay]{\node[yshift=2pt,xshift=-7pt,fill=#1,opacity=.15,fit={(A)($(B)+(.615\linewidth,.8\baselineskip)$)}] {};}\ignorespaces}
\colorlet{mypink}{red!40}
\colorlet{myblue}{cyan!60}

\newcommand{\pinkcomment}[1]{\textcolor{red!40!black}{\transparent{0.9}{\scriptsize{\texttt{\textbf{//\hspace{2pt}#1}}}}}}
\newcommand{\bluecomment}[1]{\textcolor{cyan!60!black}{\transparent{0.9}{\scriptsize{\texttt{\textbf{//\hspace{2pt}#1}}}}}}
\newcommand{\hcomment}[1]{\highlight{{\scriptsize{\texttt{\textbf{//\hspace{2pt}#1}}}}}}

\begin{algorithm}[t]
    \small{
    \caption{Thompson sampling for LLM alignment (intractable).}\label{algo:theoretical_algo}
    \hspace*{\algorithmicindent} \textbf{Input:} Prompt distribution $p_\mathcal{X}$, unknown but queryable preference oracle $\mathbb{P}$.
    \begin{algorithmic}[1]
    \State Initialize experience $\gD_0 \gets \varnothing$.
    \For{$t=1, \dots, T$}
        \State Receive a prompt $\vx_t \sim p_{\gX}$.
        \State \label{algo:line:1sty}Sample $r \sim p_r(\cdot|\gD_{t-1})$ and set $\vy_t \gets \highlight{{\arg\max}_{\vb \in \gY}}r(\vx_t, \vb)$. \hfill \hcomment{Select 1st response $\vy$.}
        \Statex \pinkcomment{E\&E objective: aligning an online system.}
        \State \tikzmk{A} \textbf{repeat} \label{algo:line:2ndyee}
        \Statex \hspace*{\algorithmicindent} \quad Sample $r \sim p_r(\cdot|\gD_{t-1})$ and set $\vy_t' \gets \highlight{\arg\max_{\vb \in \gY}}r(\vx_t, \vb)$.\hfill \hcomment{Select 2nd response $\vy'$.}
        \Statex \quad\; \textbf{until} {$\vy_t' \neq \vy_t$}
        \Statex  \tikzmk{B} \bluecomment{BAI objective: labeling via crowdsourcing.}
        \boxitee{mypink} 
        \State \label{algo:line:2ndybai}\tikzmk{A}Set $\vy_t' \gets \highlight{\arg\max_{\vb \in \gY}} \sV\left[\sigma\left(r(\vx_t, \vy_t) - r(\vx_t, \vb)\right)\right]$,\hfill \hcomment{OR select 2nd response $\vy'$.}
        \Statex \hspace*{\algorithmicindent} \quad where $\sV\left[\cdot\right]$ computes variance over the posterior $p_r(\cdot|\gD_{t-1})$.
        \State \tikzmk{B}\boxitbai{myblue} Query $\sP$ to label $\{\vy_t, \vy_t'\}$, and update experience $\gD_{t} \gets \gD_{t-1} \bigcup~\{\vx_t, \vy^+_t, \vy^-_t\}$.
    \EndFor 
    \Statex \hfill \hcomment{See \cref{algo:practical_algo} for a practical version.}
    \end{algorithmic}
    }
\end{algorithm}

As \cref{algo:theoretical_algo} presents, the first response of the duel is always selected via standard TS (\cref{algo:line:1sty}). The selection of the second response varies across different settings. \cref{algo:line:2ndyee} will be used for scenarios where preference feedback is collected from online users (the E\&E setting). The dueling responses selected in this case will both try to maximize a sampled RM, so that the online user experience is warranted with best effort. However, such algorithm can have poor asymptotic performance for BAI problems \citep{russo2016simple}, because sub-optimal responses with confidently high rewards might be tried for a long time at the expense of not exploring other potentially better choices. In light of this, \cref{algo:line:2ndybai} provides an alternative for scenarios where we could hire annotators for feedback and low-quality but exploratory responses are safe to try. Specifically, \cref{algo:line:2ndybai} selects the second response as the one that maximizes the variance of the preference (\cref{eq:bt_model}) over the first response $\vy$. This variance quantifies the \textit{epistemic uncertainty} of the RM, pointing the agent to the maximally informative direction to explore for better sample efficiency.

However, \cref{algo:theoretical_algo} is yet to be practical for LLM alignment for three main reasons. First, computing and sampling from a reward posterior is intractable for nearly all RMs at LLM scale, which are mostly based on large transformers \citep{lambert2024rewardbench}. Second, even if we managed to approximate the reward posterior, the \highlight{$\arg\max$} operations for response selection are still intractable since the search space $\gY$ is discrete and massive for token sequences of arbitrary length. Last but not least, an LLM agent~\citep{achiam2023gpt,touvron2023llama} typically consists in a generative model $\pi_\theta$ (e.g., a transformer~\citep{vaswani2017attention}), while the algorithm above is centered around a reward posterior $p(r|\gD)$ that cannot be easily converted into a generative model.

\subsection{Practical implementation}
\label{sec:sea_impl}

\subsubsection{Epistemic reward model for posterior sampling}
\label{sec:erm}

To implement active exploration with TS, we seek an efficient way to maintain and incrementally update the reward posterior $p(r|\gD)$. We consider deep ensemble for our purpose, due to its capability to model epistemic uncertainty~\citep{lakshminarayanan2017simple} and provable results when applied to TS in linear bandits~\citep{qin2022analysis}. Specifically, we update a set of plausible RMs independently and online, using the preference data and a regularized negative log-likelihood loss:
\begin{align}
    \mathcal{L}_\gR(\Phi^t|\gD_t) = \sum_{k=1}^{K} \left( \gL_r(\phi_k^t|\mathcal{D}_{t}) -  \lambda || \phi_k^t - \phi_k^0|| \right),
    \label{eq:loss_erm}
\end{align}
where $\gL_r$ is defined in \cref{eq:r_loss}, $\Phi^t = \{\phi^t_k\}_{k=1}^K$ contains the weights of the ensemble of size $K$, and $\lambda$ controls the regularization towards individual initial weights $\phi_k^0$ to retain the diversity across ensemble members~\citep{dwaracherla2024efficient}. In practice, we train $K$ MLP heads on top of a pretrained and frozen transformer. We refer to the ensemble as the Epistemic Reward Model (ERM, denoted as $\gR_\Phi$), with which the posterior sampling ($r \sim p_r(\cdot|\gD_t)$) simply amounts to randomly picking a $\phi^t_k$ from $\Phi^t$.\looseness=-1

\subsubsection{Policy-guided search to approximate \highlight{$\arg\max$}}
\label{sec:search}

With the ERM approximating the reward posterior, we need to further approximate the response selection steps (\cref{algo:line:1sty,algo:line:2ndyee,algo:line:2ndybai}) which generally take the form of $\highlight{\arg\max_{\vb \in \gY}}U(\vb)$, where $U$ absorbs the sampled prompt, the sampled RM, and optionally the selected first response (for BAI, \cref{algo:line:2ndybai}). To obtain the maximum, bandit algorithms for large action spaces typically resort to an action optimization oracle~\citep{katz2020empirical,zhu2022contextual}, but they assume a linear structure of $U$ with respect to $\vb$, which might be impractical for LLMs. Therefore, we instead replace the optimization over $\gY$ with sampling from a policy-guided distribution conditioned on $U$, $\pi_{\mathrm{prior}}(\cdot|\vx)\exp{(U(\cdot)/\eta)}$, which is appropriate since it favors responses $\vy$ that approximately maximize $U(\vy)$. In practice, for a given prompt $\vx_t$, we sample $M$ candidate responses from the prior policy $\pi_{\mathrm{prior}}(\cdot|\vx_t)$ to construct a proposal set $\gS_t = \{\vy_t^i\}_{i=1}^M$. We then conduct a greedy search in $\gS_t$ (taking $\eta \to 0$) to identify the response $\vy_t$ (or $\vy_t'$) that locally maximizes the utility function $U$, which is subsequently used in the duel. We also reuse the same $\gS_t$ for different $U$ functions at time $t$ to save computation. The choice of $\pi_{\mathrm{prior}}$ will be discussed in the next section.

\subsubsection{Online policy learning from mixed preferences}
\label{sec:mpl}
We finally resolve two remaining questions: \textit{(Q1)} how to choose a sensible $\pi_\mathrm{prior}$ at each time $t$ and \textit{(Q2)} how to get a good generative policy online. To this end, we propose a simple approach to approximately address both questions simultaneously. That is, we can utilize any direct optimizer to learn the policy $\pi_{\theta^t}$ online with the following loss and use the latest online policy as $\pi_\mathrm{prior}$:
\begin{equation}
    \label{eq:loss_pi}
    \mathcal{L}_\pi(\theta^t|\gB_t, \piref, F) = \E_{\left(\vx, \vy^+, \vy^-\right) \sim p_{\gB^t}}\left[F_{\theta^t}(\vx, \vy^+, \vy^-,\piref) \right],
\end{equation}
where $\gB_t$ is a batch of preference data labeled by the oracle wherein the responses are proposed by $\pi_{\mathrm{prior}}$ and selected by $\gR_{\Phi^t}$, $F$ could be any DAP loss (see \cref{para:dap_losses} for some examples), and $\piref$ is chosen to be $\pi_{\mathrm{sft}}$. Note that we use $\pi_{\theta^t}$ as $\pi_{\mathrm{prior}}$ at any time $t$, thus $\gB^t$ is a batch of on-policy data. By \textit{contrastive training} on these \textit{on-policy} data, we leverage their orthogonal benefits to achieve maximal policy improvement~\citep{tajwar2024preference,tang2024understanding}.

Now that optimizing \cref{eq:loss_pi} yields a good online policy $\pi_{\theta^t}$ (answering Q2), we need to assess whether $\pi_{\theta^t}$ can serve as a suitable $\pi_{\mathrm{prior}}$ for approximating the $\arg\max$ in TS (Q1). If we optimize $\pi_{\theta^t}$ with oracle preference data, $\gS_t$ will be biased towards responses with high oracle reward $r^\star$. Bias towards high-$r^\star$ region is generally helpful because it aligns with ${\arg\max_{\vb \in \gY}}r(\vx, \vb)$ that seeks high-reward responses. However, optimizing $\pi_{\theta^t}$ \textit{only} with oracle data can average out the epistemic uncertainty of $\gR$, hindering the exploration efficiency. To mitigate this issue, we further align $\pi_{\theta^t}$ with $\gR_{\Phi^t}$ using the same direct optimizer to encourage $\pi_{\theta^t}$ to propose high-$r_{\phi^t_k}$ responses for individual $r_{\phi^t_k}$, leading to better approximation of ${\arg\max_{\vb \in \gY}}r(\vx, \vb)$ for any sampled $r$. To implement, we optimize \cref{eq:loss_pi} over a mixture distribution $p_{\gB_t^{\text{mix}}} = \gamma p_{\gB_t} + (1-\gamma) p_{\gB_t^{\text{ERM}}}$, where $\gamma \in [0,1]$ controls the mixture ratio and $\gB_t^{\text{ERM}} = \{\vx_i, \Tilde{\vy}^+_i, \Tilde{\vy}^-_i\}_{i=1}^b$ consists of preference data labeled by randomly sampled individual ensemble members $r_{\phi^t_k}$. Interestingly, learning from mixed preferences further boosts sample efficiency because it utilizes the internal ERM to get pseudo labels instead of querying humans. This relates closely to model-based RL, for which we discuss further in \cref{sec:app:dyna}. We summarize our practical algorithm (\cref{algo:practical_algo}) in \cref{sec:app:algo_details}.

\vspace{-0.1cm}
\section{Experimental setup}
\vspace{-0.1cm}
\label{sec:setup}
In this section, we elaborate the experimental setup employed to validate our algorithm and ensure fair comparisons with other online alignment baselines. We start by introducing the distributed learning system designed for experimenting with online LLM alignment using simulated human preferences (\cref{sec:learning_system}). Then, we provide key experimental details in \cref{sec:experiment_details}, with a full description available in \cref{sec:app:full_details}.

\vspace{-0.1cm}
\subsection{Distributed learning system}
\vspace{-0.1cm}
\label{sec:learning_system}
The interactive nature of LLM alignment necessitates an integrated online learning system that simulates the interface depicted on the right of \cref{fig:interface}. The absence of a performant open-source online alignment system has restricted many existing works to only a few iterations of batch learning~\citep{muldrew2024active,dong2024rlhf,chen2024bootstrapping,zhang2024self,xie2024exploratory}, which creates a mismatch with their theories that typically require a large number of online interaction rounds. Even worse, such absence also makes the comparison between different LLM exploration methods difficult, often restricting evaluations to the simplest iterative DAP baselines~\citep{zhang2024self,xie2024exploratory}.

\begin{wrapfigure}{R}{0.3\textwidth}
    \vspace{-1.7em}
    \begin{minipage}{0.3\textwidth}
    \includegraphics[width=\textwidth]{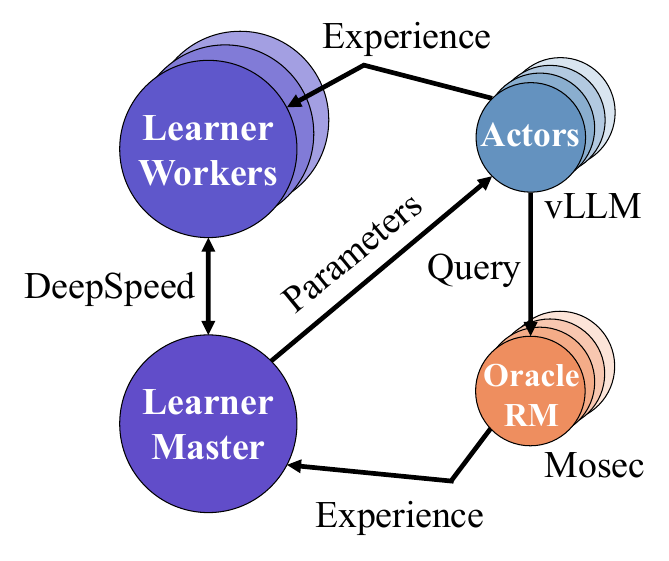}
    \caption{The learning system for experimenting online LLM alignment algorithms.}
    \label{fig:system}
    \vspace{-0.9em}
    \end{minipage}
\end{wrapfigure}

To fill this gap, we build a highly efficient learning system for experimenting with online LLM alignment algorithms. We notice that the computational bottleneck lies in online response sampling (i.e., autoregressive generation) and preference labeling (e.g., human, large RMs, or large LLMs), which mirrors the slow actor-environment interaction seen in RL systems. Inspired by distributed deep RL systems which spawn many actors or environments in parallel~\citep{espeholt2018impala,weng2022envpool}, we design an Actor-Learner-Oracle architecture for online LLM alignment, which is depicted in \cref{fig:system}. The three types of workloads (i.e., actor, learner and oracle) are heterogeneous and require different optimization. In particular, we adopt vLLM~\citep{kwon2023efficient} for the actor to accelerate the autoregressive response generation. We also use DeepSpeed's ZeRO~\citep{rasley2020deepspeed,rajbhandari2020zero} strategies to enhance the memory efficiency of the learner. The updated model weights are broadcasted from the learner master to all actors after every optimizer step efficiently via NCCL, similar to \citet{hu2024openrlhf}. Furthermore, to improve the scalability, we wrap the oracle RM as a service using Mosec~\citep{yang2021mosec}, which supports dynamic batching and parallel processing, to minimize preference query latency. Finally, we leverage DeepMind Launchpad~\citep{yang2021launchpad} to compose all workloads into a distributed program and adopt Plasma~\citep{moritz2017plasma} to efficiently transfer data across process boundaries.

We benchmark our system's efficiency against a concurrent implementation of online DPO by HuggingFace\footnote{\url{https://huggingface.co/docs/trl/main/en/online_dpo_trainer}.}, which utilizes only DeepSpeed for memory optimization. Our system achieves up to \highlight{$\textbf{2.5}\times$} latency reduction compared to this counterpart, demonstrating its computational efficiency. Due to space constraints, detailed benchmarking methods and results are presented in \cref{sec:app:benchmark}. Our codebase, \textbf{oat} (\underline{o}nline \underline{a}lignmen\underline{t}), along with the implementation of \sea, is open-sourced at {\url{https://github.com/sail-sg/oat}} to accelerate future research in online LLM alignment.

\subsection{Experiment details}
\label{sec:experiment_details}
We adopt SFT models tuned on \texttt{TL;DR}~\citep{stiennon2020learning} from~\citet{huang2024n+}, which cover three scales (1B, 2.8B, 6.9B) of the Pythia family~\citep{biderman2023pythia}, as starting points for our experiments. We choose~\citet{skyworkreward2024} to be the oracle RM.  To verify the effectiveness of~\sea, we employ three direct optimizers: DPO~\citep{rafailov2023dpo}, IPO~\citep{azar2024ipo}, and SLiC~\citep{zhao2023slic}. Besides, two LLM exploration methods built on DPO, APL~\citep{muldrew2024active} and XPO~\citep{xie2024exploratory}, are fairly compared when using DPO as the optimizer. Our experiments primarily focus on the BAI setting (crowdsourcing labeling), and we use win rate against reference responses as the metric. We refer readers to \cref{sec:app:full_details} for more details.

\newcommand{\offline}{{Offline}}
\newcommand{\online}{{Online}}
\newcommand{\xpo}{{XPO}}
\newcommand{\apl}{{APL}}

\newcommand{\vvone}{{Variant-1}}
\newcommand{\vvtwo}{{Variant-2}}
\newcommand{\vvthree}{{Variant-3}}
\newcommand{\vvfour}{{Variant-4}}
\newcommand{\vvfive}{{Variant-5}}
\newcommand{\vvsix}{{Variant-6}}
\newcommand{\vvseven}{{Variant-7}}

\newcommand{\expuct}{{Uncertainty}}
\newcommand{\expee}{{E\&E-TS}}
\newcommand{\expbai}{{BAI-TS}}

\section{Empirical results}
\label{sec:empirical_results}
In this section, we present our empirical results and analyses, organized into four parts: (1)~an overall comparison between \sea~and baselines across various direct optimizers and model scales; (2)~an ablation analysis to study the effects of \sea{}'s key components; (3)~a comparison of different exploration strategies under E\&E and BAI settings; (4)~additional results for alignment with a human oracle simulated by GPT4o-mini.

\begin{figure}[t]
    \centering
    \includegraphics[width=0.9\linewidth]{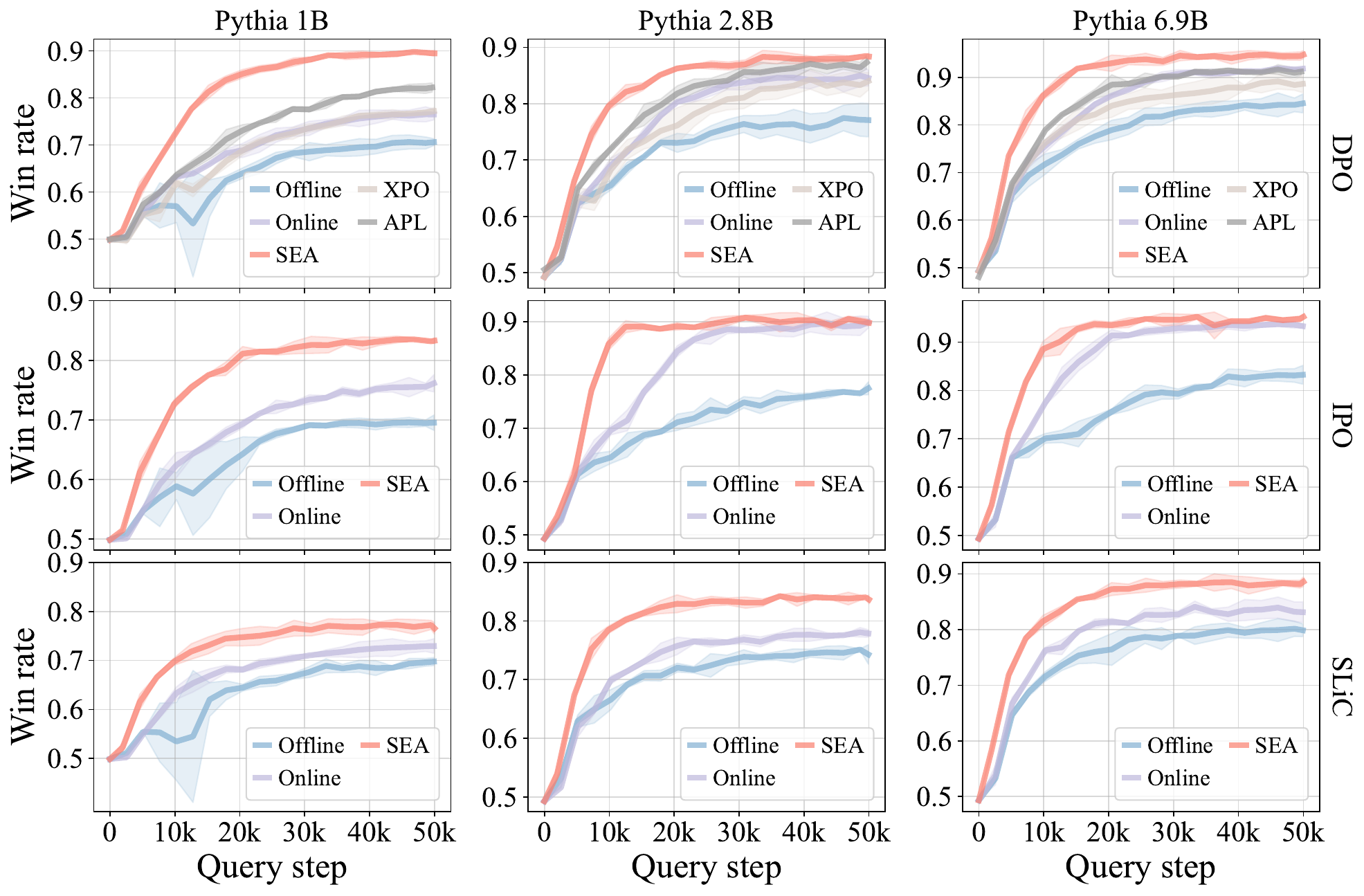}
    \caption{Win rate comparison of different algorithms against their initial SFT models across three scales and three direct optimizers.}
    \label{fig:main-results}
\end{figure}

\subsection{Overall comparison}
\label{sec:overall_cmp}
We first compare \sea{} with all baselines across three model scales and three direct optimizers. APL and XPO are only compared when DPO is used as the direct optimizer, because they are incompatible with IPO or SLiC. \cref{fig:main-results} shows the win rate curves versus the number of query steps. Across all settings, Online agents consistently improve sample efficiency over their Offline counterparts, validating the necessity of \cref{property1} for alignment algorithms. Focusing on the first row, we observe that among prior active exploration methods, XPO gives a small improvement in final performance over Online (passive) at the 1B scale, but falls short for larger scales. On the other hand, \apl{} shows a significant sample efficiency boost at the 1B scale, but this advantage diminishes when scaling up and it performs almost the same as \online{} at 6.9B scale. Our method, \sea, outperforms both offline and online passive methods across all scales and all direct optimizers, confirming the critical role that \cref{property2} plays for sample-efficient alignment. Meanwhile, in the special case of using DPO as the direct optimizer, \sea{} also shows superior performance to prior online active exploration methods including \apl{} and \xpo. We invite readers to revisit \cref{fig:3-scale-benchmark}, where we show that \sea{} not only attains significantly improved final performance (Left) but also achieves $2\text{-}5\times$ better sample efficiency (Right).\looseness=-1

Additionally, we note that the choice of direct optimizer matters for both online learning and active exploration. When comparing different optimizers at 1B scale (the first column), all \offline{} agents demonstrate comparable learning efficiency and reach the same level of final performance (around $70\%$ win rate), but SLiC \online{} agent deliver slightly less improvement than DPO and IPO \online{} agents. Besides, when incorporating active exploration, the \sea{} agent using DPO shows much larger improvement than the other two. This suggests that selecting the most suitable policy optimizer coupled with active exploration would yield the best agent.\looseness=-1

\subsection{Ablation analysis}

\begin{table}[t]
    \centering
    \small{
    \caption{Decomposition of different driving factors of online active alignment algorithms.}
    \begin{tabular}{ccccl}
         \toprule
         Variant & Inference (Test) & Exploration & Learn & Remark \\
         \toprule
         \toprule
         1 & $\pi_\theta$ & passive & $\pi_\theta$ & Online \texttt{DAP}~\citep{guo2024direct} \\
         2 & $\pi_\theta$ & active & $(\pi_\theta, \gR_\Phi)$ & \sea~\textit{without} ERM sync (\cref{sec:mpl}) \\
         3 & $\pi_\theta$ & active & $(\pi_\theta \leftrightarrow \gR_\Phi)$ & \sea \\
         \midrule
         4 & $\operatorname{BoN}(\pi_\theta, \gR_\Phi)$ & passive & $(\pi_\theta, \gR_\Phi)$ & - \\
         5 & $\operatorname{BoN}(\pi_\theta, \gR_\Phi)$ & active & $(\pi_\theta, \gR_\Phi)$ & - \\
         6 & $\operatorname{BoN}(\pi_\theta, \gR_\Phi)$ & active & $(\pi_\theta \leftrightarrow \gR_\Phi)$ & \sea~with Best-of-N sampling \\
         \midrule
         7 & $\operatorname{BoN}(\piref, \gR_\Phi)$ & active & $\gR_\Phi$ & Not learn policy~\citep{dwaracherla2024efficient} \\
         \bottomrule
    \end{tabular}
    \label{tab:analysis}
    }
\end{table}

\begin{wrapfigure}{R}{0.6\linewidth}
    \vspace*{-0.7cm}
    \centering
    \includegraphics[width=\linewidth]{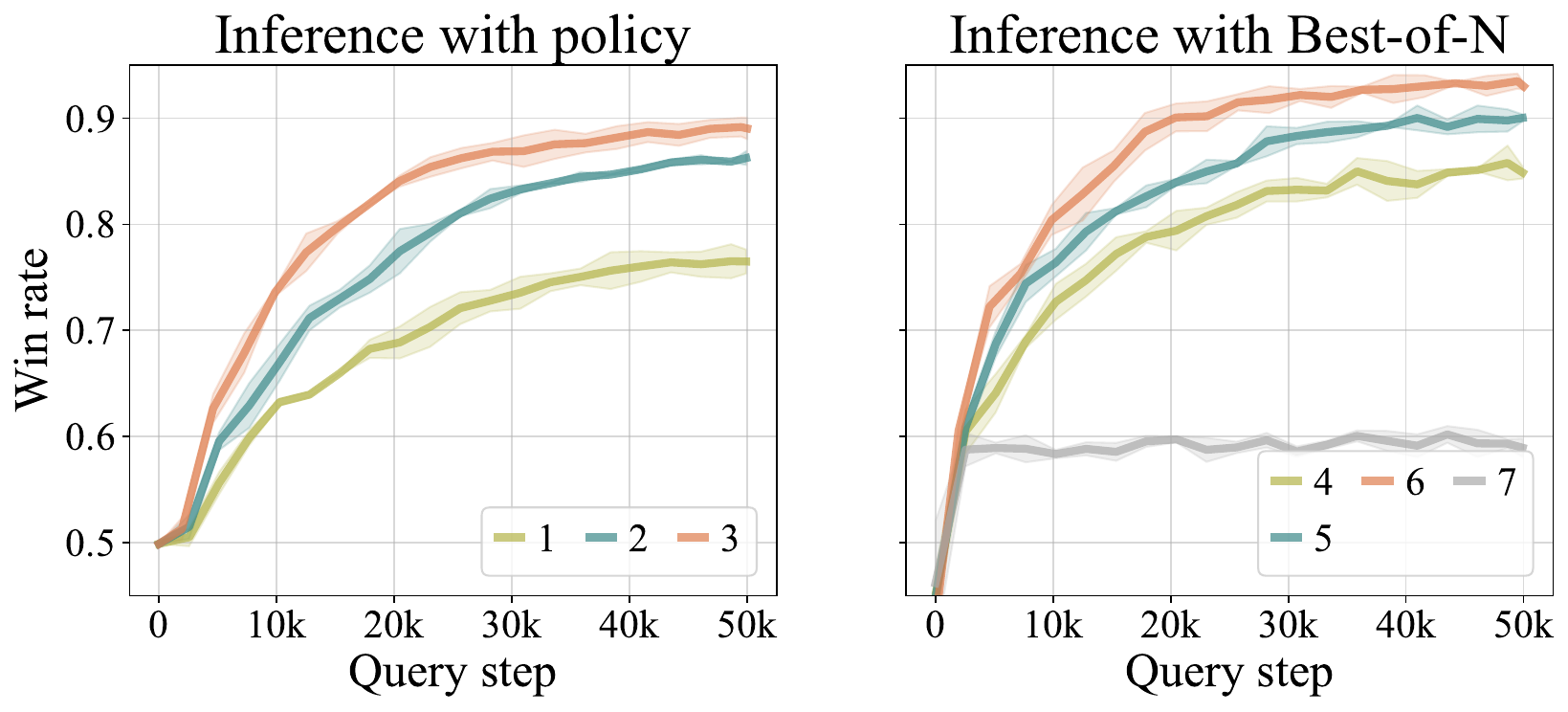}
    \caption{Win rate comparison of different agent variants when using \textbf{(Left)} policy and \textbf{(Right)} Best-of-N sampling for inference.}
    \label{fig:analysis}
\end{wrapfigure}

Next, we decompose \sea{} into distinct components to evaluate their individual contributions. \cref{tab:analysis} shows the three axes we dissect \sea{} on, including inference methods, exploration strategies, and learning components. We construct seven agent variants from different combinations, which cover two closely related baselines~\citep{guo2024direct,dwaracherla2024efficient}. We show in \cref{fig:analysis} the performance curves of each variant, all trained with DPO on 1B scale. The left plot compares variants that directly use the policy for inference. It clearly shows the benefits of learning ERM for active exploration (\vvtwo) and aligning $\pi_{\theta^t}$ with $\gR_{\Phi^t}$ (\vvthree). Since a reward model is learned within the agent, we can further incorporate inference-time alignment via Best-of-N (BoN) sampling~\citep{nakano2021webgpt,touvron2023llama}. This also facilitates a direct comparison between \sea{} and \citet{dwaracherla2024efficient}, which learns a similar ERM for both exploration and BoN but does not align the LLM policy. Results in the right plot of \cref{fig:analysis} suggest a similar trend that \vvsix$~\succ~$\vvfive$~\succ~$\vvfour. The \vvseven~\citep{dwaracherla2024efficient}, however, ceases to improve after ERM converges due to the limited performance of its fixed policy.

\subsection{Choice of exploration strategies}
\label{sec:choice_of_exploration}

\begin{figure}[t]
    \centering
    \includegraphics[width=0.9\linewidth]{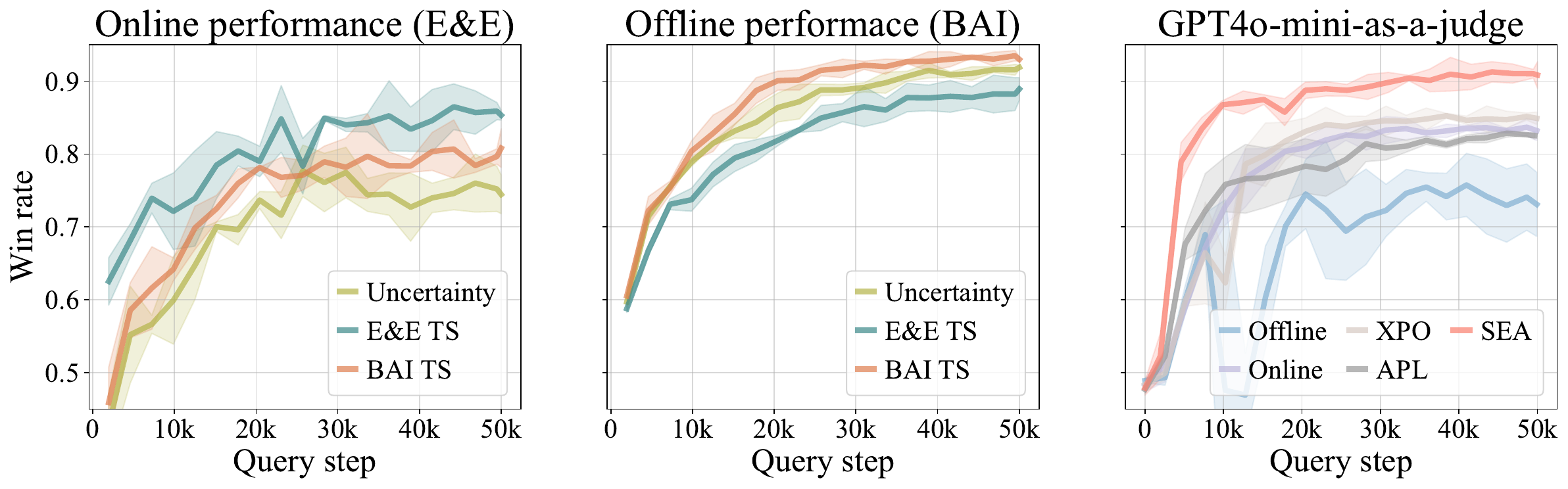}
    \vspace{-.2cm}
    \caption{\textbf{(Left and Middle)} Win rate comparison of different exploration strategies measured in E\&E and BAI settings. \textbf{(Right)} Win rate comparison of different agents when using \texttt{GPT4o-mini} to simulate human feedback via LLM-as-a-judge.}
    \vspace{-.6cm}
    \label{fig:gpt_exp}
\end{figure}

Recalling that different LLM alignment scenarios (online system or crowdsourcing) require different exploration strategies to meet their respective learning objectives (\cref{para:different_obj}). We investigate three strategies based on posterior sampling and compare them on both online and offline performance. The first strategy (\expuct) focuses on pure exploration with information maximization. It seeks the pair of dueling responses that exhibits the largest epistemic uncertainty, which is implemented by selecting the pair whose logits difference has the largest variance across ensemble members. The second (\expee) and the third (\expbai) strategies follow the principles in \cref{algo:theoretical_algo}, and their differences are between \cref{algo:line:2ndyee} and \cref{algo:line:2ndybai}. The comparison results are shown in \cref{fig:gpt_exp} (Left and Middle). Focusing on the left plot, we observe that \expee{} strategy achieves the best online performance, which is within our expectation. In contrast, \expuct{} shows the worst online performance because it tries to maximize the information gain but does not prioritize reward maximization. On the other hand, conclusions are interestingly different when taking the offline performance as the metric. In this case, \expbai{} and \expuct{} both exhibit more efficient offline performance improvement than \expee. This can be attributed to that exploration for uncertainty minimizing helps to identify more informative responses to train the LLM policy. Moreover, \expbai~$\succ$~\expuct{} indicates exploration with both reward and information maximization is better than exploration with only information maximization. \expee, however, always chooses two responses with similarly high quality to exploit. This can not only lead to less efficient exploration, but also result in less efficient policy learning due to smaller DAP loss gradients.\looseness=-1

\vspace{-0.1cm}
\subsection{Aligning LLMs with a human simulator}
\vspace{-0.1cm}
Results presented so far are based on experimenting LLM alignment with the preference oracle being a scalar reward model, which is deterministic and does not capture the potential randomness of the choice by real humans. To test different agents in a more realistic setting, we use generative models as human simulator in an LLM-as-a-judge~\citep{bubeck2023sparks,zheng2023judging} manner. In particular, we directly query the OpenAI API and use the \texttt{gpt-4o-mini-2024-07-18} model as the judge to provide preference feedback. We follow the prompt template of~\cite{alpaca_eval}. The results are shown in \cref{fig:gpt_exp} (Right). We can observe the performance curves generally exhibit higher variance, possibly due to the randomness introduced in the feedback process, which puts more stringent requirements for learning algorithms. The two active exploration methods demonstrate opposite results to those in \cref{sec:overall_cmp}---\apl{} learns fast initially but is eventually outperformed by \online, while~\xpo~improves over~\online~after stabilizing its training and delivers a better final performance. Our agent, \sea, is shown to offer the best sample efficiency as well as asymptotic performance, further validating the importance of online learning and well-designed active exploration mechanism.\looseness=-1

\vspace{-0.1cm}
\section{Conclusion}
\vspace{-0.1cm}
In this paper, we study the problem of LLM alignment through the lens of contextual dueling bandits and propose a Thompson sampling-based algorithm to achieve sample-efficient alignment. We incorporate three techniques, including epistemic reward model, policy-guided search and mixed preference learning to yield a practically efficient online alignment method. Extensive empirical evaluation demonstrates the superior sample efficiency of our method compared to existing baselines. To our knowledge, this is the first work to study active exploration for online LLM alignment with fully online experimental verification. We hope our positive empirical results, along with the open-sourced codebase, will encourage future research in this direction, ultimately enabling LLMs to achieve superhuman intelligence with an affordable amount of human feedback.
\small{
\bibliography{reference}
\bibliographystyle{iclr2025_conference}
}
\newpage

\appendix

\section{Algorithm details}
\label{sec:app:algo_details}
While \cref{algo:theoretical_algo} presents our Thompson sampling algorithm for LLM alignment, it is intractable and centered around the reward posterior modeling. We next present a practical sample-efficient alignment agent that learns both an LLM policy and an epistemic reward model online in \cref{algo:practical_algo}.\looseness=-1

\newcommand{\boxiteetwo}[1]{\tikz[remember picture,overlay]{\node[yshift=2pt,xshift=-7pt,fill=#1,opacity=.15,fit={(A)($(B)+(.73\linewidth,.8\baselineskip)$)}] {};}\ignorespaces}
\newcommand{\boxitbaitwo}[1]{\tikz[remember picture,overlay]{\node[yshift=2pt,xshift=-7pt,fill=#1,opacity=.15,fit={(A)($(B)+(.695\linewidth,.8\baselineskip)$)}] {};}\ignorespaces}

\begin{algorithm}[h]
    \small{
    \caption{Sample-efficient alignment (\sea) for LLMs}\label{algo:practical_algo}
    \hspace*{\algorithmicindent} \textbf{Input:} Reference policy $\piref$, DAP loss function $F$, prompt distribution $p_\mathcal{X}$, unknown but queryable preference oracle $\mathbb{P}$, mixture ratio $\gamma$.
    \begin{algorithmic}[1]
    \State Initialize experience $\gD_0 \gets \varnothing$, policy $\pi_{\theta^0} \gets \piref$, and ERM weights $\Phi^0 = \{\phi_k^0\}_{k=1}^K$ randomly.
    \For{$t=1, \dots, T$}
        \State Receive a prompt $\vx_t \sim p_{\gX}$.
        \State Sample $M$ responses $\vy_t^i \sim \pi_{\theta^{t-1}}(\cdot|\vx_t)$ to construct $\gS_t = \{\vy_t^i\}_{i=1}^M$.
        \State Sample $\phi \sim \operatorname{Uniform}(\Phi^{t-1})$ and set $\vy_t \gets {\arg\max}_{\vb \in \gS_t}r_{\phi}(\vx_t, \vb)$. \hfill \hcomment{Select 1st response $\vy$.}
        \Statex \pinkcomment{E\&E objective: aligning an online system.}
        \State \tikzmk{A} \textbf{repeat}
        \Statex \hspace*{\algorithmicindent} \quad Sample $\phi \sim \operatorname{Uniform}(\Phi^{t-1})$ and set $\vy_t' \gets {\arg\max}_{\vb \in \gS_t}r_{\phi}(\vx_t, \vb)$.\hfill \hcomment{Select 2nd response $\vy'$.}
        \Statex \quad\; \textbf{until} {$\vy_t' \neq \vy_t$}
        \Statex  \tikzmk{B} \bluecomment{BAI objective: labeling via crowdsourcing.}
        \boxiteetwo{mypink} 
        \State \tikzmk{A}Set $\vy_t' \gets \arg\max_{\vb \in \gS_t} \sV_\phi\left[\sigma\left(r_\phi(\vx_t, \vy_t) - r_\phi(\vx_t, \vb)\right)\right]$,\hfill \hcomment{OR select 2nd response $\vy'$.}
        \Statex \hspace*{\algorithmicindent} \quad where $\sV_\phi\left[\cdot\right]$ computes variance across ensemble members of $\Phi^{t-1}$.
        \State \tikzmk{B}\boxitbaitwo{myblue} \textbf{if} $g < \gamma$ for $g \sim \operatorname{Uniform(0,1)}$ \textbf{then}
        \Statex \hspace*{\algorithmicindent} \quad Label $\{\vy_t, \vy_t'\}$ with $\sP$ to obtain $\gB_t = \{\vx_t, \vy^+_t, \vy^-_t\}$ and update experience $\gD_{t} \gets \gD_{t-1} \bigcup \gB_t$.
        \Statex \quad\; \textbf{else}
        \Statex \hspace*{\algorithmicindent} \quad Use $\gR_{\Phi^{t-1}}$ to get synthetic labels and obtain $\gB_t = \{\vx_i, \Tilde{\vy}^+_i, \Tilde{\vy}^-_i\}$.
        \Statex \quad\; \textbf{end if}
        \State Update ERM with the regularized NLL loss (\cref{eq:loss_erm}):
        \Statex $$ \Phi^t \gets \Phi^{t-1} - \alpha_\gR \nabla_\Phi \gL_{\gR}(\Phi^{t-1} | \gD_t).$$ \hfill \hcomment{Reward learning.}
        \State Update policy with the direct optimizer (\cref{eq:loss_pi}):
        \Statex $$ \theta^t \gets \theta^{t-1} - \alpha_\pi \nabla_\theta \gL_{\pi}(\theta^{t-1} | \gB_t, \piref, F).$$ \hfill \hcomment{Policy learning.}
    \EndFor
    \end{algorithmic}
    }
\end{algorithm}
In \cref{algo:practical_algo}, we describe an online setting where a single example is processed at each time $t$ (batch size $b=1$). This is mainly for notational convenience, while in implementation we set $b$ to be the training batch size (e.g., $128$). We instantiate the reward posterior with an epistemic reward model, which allows for efficient incremental update and sampling. We also replace the global optimization ($\arg\max_{\vb \in \gY}$) with a policy-guided local search among proposals sampled from the latest online policy $\pi_{\theta^{t-1}}$. At each time $t$, we update ERM weights $\Phi$ with $m$ randomly sampled batches from the experience $\gD_t$. We find setting $m=5$ suffices to achieve reasonable accuracy. The policy parameters $\theta$ are updated using mixed preference data, with proportion $\gamma$ being the real environment experience and $(1-\gamma)$ from the ERM's synthetic experience. Note that the synthetic experience is not added into $\gD_t$ to ensure reward learning always uses ground truth environment data.\looseness=-1

\label{para:dap_losses}We consider the following three direct optimizers in our experiments:
\begin{itemize}
    \item DPO~\citep{rafailov2023dpo}: \begin{equation}F_{\theta}(\vx, \vy^+, \vy^-,\piref) = -\log \sigma\left(\beta \log \frac{\pi_{{\theta}}\left(\boldsymbol{y}^{+} | \boldsymbol{x}\right) \pi_{\mathrm{ref}}\left(\boldsymbol{y}^{-} | \boldsymbol{x}\right)}{\pi_{\mathrm{ref}}\left(\boldsymbol{y}^{+} | \boldsymbol{x}\right) \pi_{{\theta}}\left(\boldsymbol{y}^{-} | \boldsymbol{x}\right)}\right)\end{equation}
    \item IPO~\citep{azar2024ipo}: \begin{equation}F_{\theta}(\vx, \vy^+, \vy^-,\piref) = \left( \log \left( \frac{\pi_{{\theta}}\left(\boldsymbol{y}^{+} | \boldsymbol{x}\right) \pi_{\mathrm{ref}}\left(\boldsymbol{y}^{-} | \boldsymbol{x}\right)}{\pi_{\mathrm{ref}}\left(\boldsymbol{y}^{+} | \boldsymbol{x}\right) \pi_{{\theta}}\left(\boldsymbol{y}^{-} | \boldsymbol{x}\right)}\right) - \frac{1}{2\beta} \right)^2\end{equation}
    \item SLiC~\citep{zhao2023slic}: \begin{equation}F_{\theta}(\vx, \vy^+, \vy^-,\piref) = \max \left(0, 1 - \beta \log \frac{\pi_{{\theta}}\left(\boldsymbol{y}^{+} | \boldsymbol{x}\right) \pi_{\mathrm{ref}}\left(\boldsymbol{y}^{-} | \boldsymbol{x}\right)}{\pi_{\mathrm{ref}}\left(\boldsymbol{y}^{+} | \boldsymbol{x}\right) \pi_{{\theta}}\left(\boldsymbol{y}^{-} | \boldsymbol{x}\right)}\right)\end{equation}
\end{itemize}
where $\beta$ controls the rate of deviation of $\pi_\theta$ from $\piref$.\looseness=-1

\section{On connections with single-step RL}
\label{sec:app:dyna}
By viewing contextual dueling bandits as \textit{single-step} preference-based RL (PbRL)~\citep{busa2014preference,wirth2017survey} problems, we can interpret paradigms shown in \cref{fig:compare} from the RL perspective.\looseness=-1

RLHF approaches (\cref{fig:compare}{\color{Refcolor}a}) are instances of \textbf{offline model-based RL}~\citep{kidambi2020morel,yu2021combo,schrittwieser2021online,liu2023rosmo,tajwar2024preference}, where they learn a reward model (no need for a transition model since the prompt-response interaction is single-step) of the environment from a batch of offline collected data, and train a policy (i.e., LLM) to maximize the return (i.e., expected one-step reward) with respect to the \textit{learned} reward.\looseness=-1

In contrast, DAP methods (\cref{fig:compare}{\color{Refcolor}b}) are similar to \textbf{policy-based model-free RL} algorithms, e.g., REINFORCE~\citep{williams1992simple} which conducts policy gradient update:
\begin{equation}
    \label{eq:reinforce}
    \E_{\vx \sim \gX} \E_{\vy \sim \pi_\theta(\cdot | \vx)} \left[ R(\vx, \vy) \nabla_\theta \log \pi_\theta(\vy|\vx) \right],
\end{equation}
where $R(\vx, \vy)$ is the return (i.e., cumulative reward) of the trajectory. To connect with DAP, we could set $R$ as arbitrary scalar values based on the binary preference outcomes, e.g., $R(\vx, \vy^+) = \zeta$ and $R(\vx, \vy^-) = -\zeta$ for preference triplet $\{\vx, \vy^+, \vy^-\}$. In this way we could rewrite \cref{eq:reinforce} as
\begin{equation}
    \E_{\vx \sim \gX} \E_{\vy,\vy' \sim \pi_\theta(\cdot | \vx)} \E_{(\vy^+ \succ \vy^-) \sim \sP} \left[ \zeta \left(\nabla_\theta \log \pi_\theta(\vy^+|\vx) - \nabla_\theta \log \pi_\theta(\vy^-|\vx) \right)\right],
\end{equation}
by repeating action sampling twice and querying the oracle for preference labeling. This matches the gradient direction of contrastive DAP losses (e.g., see Section 4 of DPO~\citep{rafailov2023dpo}) if we optimize them online~\citep{guo2024direct}.

Additionally, active reward learning from behavior policy's data distribution (\cref{fig:compare}{\color{Refcolor}c}) can be regarded as \textbf{inverse RL}~\citep{ng2000algorithms}, which tries to recover environment's reward function given expert trajectories. In the context of LLM alignment, the preference data $\{\vx, \vy^+, \vy^-\}_{i=1}^N$ directly encodes human's implicit reward $r^\star$, which can be inversely learned with assumptions such as the BT model~\citep{bradleyterry1952rank}. However, existing methods belonging to this paradigm mostly rely on a fixed (and suboptimal) behavior policy for response sampling, whose coverage inherently limits the quality of the recovered reward function.\looseness=-1

Last but not least, \sea~depicted in \cref{fig:compare}{\color{Refcolor}d} resembles a class of \textbf{online model-based RL} algorithms, known as Dyna~\citep{sutton1990dyna,janner2019trust}, that learns a \textit{world model} from environment experience and trains a base agent (consisting of reactive policies and value functions) from both environment experience and model experience. Compared to model-free methods, Dyna naturally enables more sample-efficient learning by planning with the learned world model to update the base agent. In~\sea, we learn the reward model online and update the LLM (i.e., the reactive policy) with model-planing experience by mixed preference learning (\cref{sec:mpl}). Online model-based RL algorithms could suffer from catastrophic forgetting in the face of nonstationary data~\citep{liu2024losse}, and we leave it for future work. Overall, this model-based RL formulation is powerful and explains popular LLM techniques, e.g., Best-of-N sampling~\citep{touvron2023llama} can be viewed as planning for acting, which trades compute for performance. We believe it is a promising path leading us to unlock superhuman capabilities of LLMs.\looseness=-1

\section{System benchmarking}
\label{sec:app:benchmark}
We conduct a rigorous benchmarking comparison on the efficiency of online DPO training using our learning system \texttt{oat}\footnote{\url{https://github.com/sail-sg/oat}.}, alongside the \texttt{trl}'s implementation\footnote{\url{https://github.com/huggingface/trl/blob/main/trl/trainer/online_dpo_trainer.py}.}.

\textbf{Settings}. In alignment with the examples provided by \texttt{trl}, we use the {TL;DR}~\citep{stiennon2020learning} dataset and evaluate training efficiency at three model scales: 1B, 2.8B and 6.9B parameters for both SFT-ed LLMs and exclusively trained RMs. This is similar to the settings in our experiments (see \cref{sec:app:full_details}) except that we fix the reward oracle to be a strong general-purpose RM.

\begin{figure}[t]
    \centering
    \includegraphics[width=\linewidth]{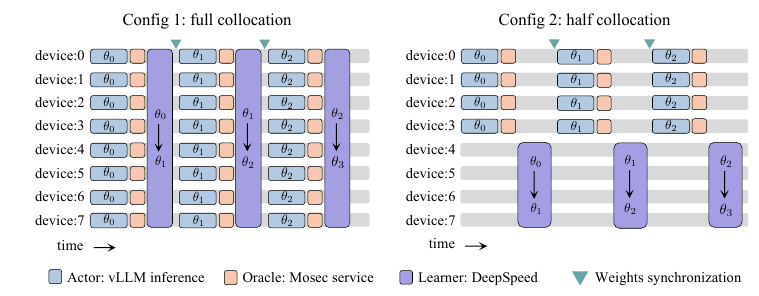}
    \caption{Two example configurations of \texttt{oat} used in benchmarking experiments.}
    \label{fig:system_configs}
\end{figure}

\textbf{Hardware \& Software}. All benchmarking experiments are conducted on a single machine with eight A100-40G GPUs and $96$ AMD EPYC 7352 CPUs. To ensure fair comparison, we align all key hyperparameters for both \texttt{oat} and \texttt{trl}. The DeepSpeed ZeRO-2 strategy is employed by default when GPU memory suffices; otherwise, ZeRO-3 or ZeRO-2-offload is utilized as applicable. Notably, the distributed architecture of \texttt{oat} provides flexibility in system configuration, enabling adjustments to accommodate memory and computational time constraints. \cref{fig:system_configs} illustrates two example configurations employed in our benchmarking experiments. 
\begin{itemize}
    \item \textbf{Config 1} collocates all three workloads on each of the GPUs. Specifically, eight vLLM instances (for actors) and eight Mosec workers (for oracle RMs) are spawned to run independently on each GPU. After a batch of responses is generated (by actors) and labeled (by oracle RMs), it is sent to the learner, which runs on all eight GPUs coordinated through ZeRO strategies for policy learning. The updated policy weights are then broadcasted to all actors for \textit{on-policy} response sampling on subsequent prompt batch. While this configuration maximizes GPU utilization, it requires substantial GPU memory to accommodate all workloads and is thus employed only for 1B scale experiments.
    \item \textbf{Config 2} only collocates actor and oracle workloads on half of the GPUs, reserving the remaining four GPUs exclusively for the learner. This is suited for larger-scale experiments (e.g., 2.8B or 6.9B), where additional GPU memory is allocated to the learner. However, this setup incurs idle time on half of the GPUs due to data dependency, as the learner must await new preference data, and the actor must await updated policies. An alternative is to implement \textit{asynchronous} data collection, where minor data staleness is allowed by using $\theta_{t-1}$ to generate data for updating $\theta_{t+1}$. Although this data would not be strictly on-policy, asynchronous training could reduce idle time and enhance GPU utilization. This approach has proven effective in large-scale RL systems~\citep{berner2019dota}, and we leave this optimization to future work.
\end{itemize}
We provide all benchmarking scripts in our codebase\footnote{\url{https://github.com/sail-sg/oat/tree/main/benchmark}.} for reproducibility.

\begin{figure}[h]
    \centering
    \includegraphics[width=0.8\linewidth]{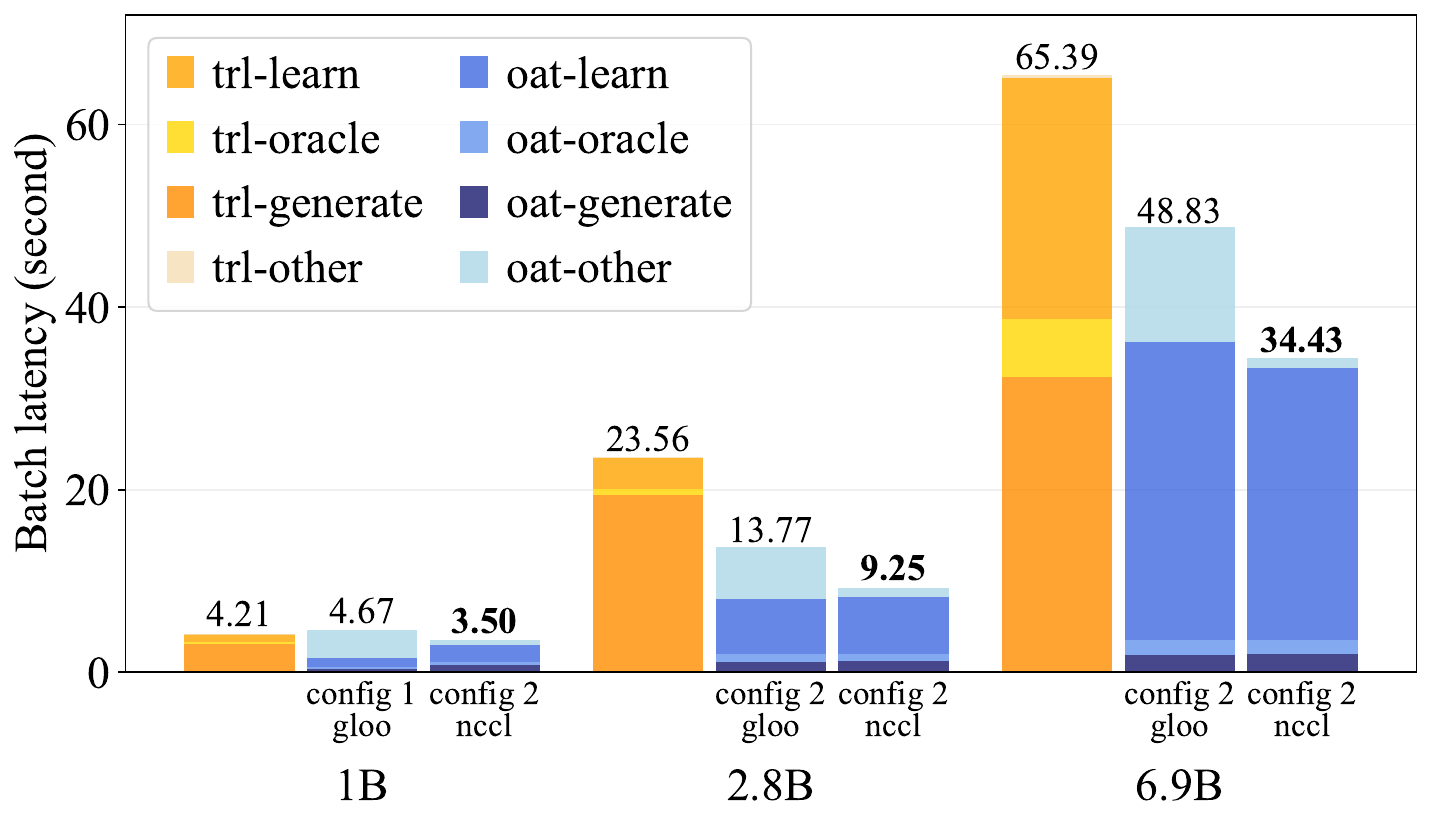}
    \caption{Averaged training latency (over 10 batches, equivalent to 1280 samples) comparing \texttt{sail-sg/oat} against \texttt{huggingface/trl}.}
    \label{fig:bench_results}
\end{figure}

\textbf{Results}. Benchmarking results for the latency of training a batch of 128 samples are presented in \cref{fig:bench_results}. Overall, training with \texttt{oat} config 2 demonstrates consistently greater efficiency than \texttt{trl}, achieving up to a \textbf{2.5}$\times$ reduction in latency at the 2.8B scale.

We next analyze the time costs for individual stages: generate, oracle and learn. Across all scales and configurations, \texttt{oat} demonstrates significantly lower \textit{generate} time than \texttt{trl}, due to distributed actors utilizing vLLM. Additionally, at the 6.9B scale, \texttt{oat} requires substantially less \textit{oracle} time than \texttt{trl}, as \texttt{trl} employs ZeRO-3 to prevent GPU memory overflow, thereby slowing inference. In contrast, \texttt{oat} config 2 allows for flexible collocation, enabling oracle RMs hosted via Mosec to operate in parallel without sharding. However, \texttt{oat} config 2 incurs longer \textit{learn} time compared to \texttt{trl} due to the use of only half the available GPUs. This limitation also explains why, at the 1B scale, config 2 has higher latency than config 1 across all stages.

The \textit{other} category accounts for time costs associated with data loading, tokenization, and communication. Here, inter-process communication is the primary cost, with \texttt{trl} showing minimal overhead as all three stages operate within the same process on identical micro-batches, avoiding weight synchronization. By contrast, \texttt{oat} requires considerable time to transfer updated policy weights from the learner to all actors. While NCCL is recommended for synchronization over GLOO, it requires older vLLM packages (prior to version 0.4.3), which may lack support for newer LLM architectures. Moreover, NCCL is incompatible with config 1 due to its restriction on the learner master process establishing two separate process groups (one for DeepSpeed, the other for weight synchronization). In summary, we recommend future researchers prioritize \texttt{oat} config 2 and employ NCCL when feasible.

\section{Full experimental details}
\label{sec:app:full_details}

\textbf{Models}. We experiment three model scales (1B, 2.8B, 6.9B) from the Pythia family~\citep{biderman2023pythia}. We take pretrained SFT models from~\cite{huang2024n+} as $\piref$ for the starting model in all experiments. Except in \cref{sec:overall_cmp}, we use 1B model for experiments to save computation.\looseness=-1

\textbf{Reward oracle}. We simulate the process of human feedback with a strong scalar RM and refer it as reward oracle. We choose \texttt{Skywork-Reward-Llama-3.1-8B}\footnote{\url{https://huggingface.co/Skywork/Skywork-Reward-Llama-3.1-8B.}}~\citep{skyworkreward2024}, which is top-ranked in RewardBench leaderboard~\citep{lambert2024rewardbench}, as the reward oracle.\looseness=-1

\textbf{Epistemic reward model}. We build ERM on top of a pretrained $0.4$B transformer~\citep{llm-blender-2023}, by removing its head and adding an ensemble of MLPs. The size of ensemble is set to $K=20$, and all MLPs contain 2 hidden layers of 128 nodes. Note that the ERM is chosen to be much smaller than the reward oracle following~\citet{dwaracherla2024efficient}, which reflects the fact that human preferences can be more complex than what the agent can model. The regularization coefficient $\lambda$ is fixed to be $0.5$ after a coarse hyperparameter search.\looseness=-1

\textbf{Data}. We employ the widely adopted \texttt{TL;DR} dataset~\citep{stiennon2020learning} for our experiments. It consists of Reddit posts as prompts, and the agent is required to give summaries that align with human preferences. We fix $50$k prompts for training and limit the query budget to $50$k as well.\looseness=-1

\textbf{DAP methods}. We adopt three DAP methods (direct optimizers) to thoroughly validate our algorithm, including DPO~\citep{rafailov2023dpo}, IPO~\citep{azar2024ipo} and SLiC~\citep{zhao2023slic}.\looseness=-1

\textbf{Baselines}. We include the offline and online variants of different DAP methods as baselines, which are studied by~\citet{guo2024direct}. Additionally, we compare with two active exploration baselines built on online DPO: APL~\citep{muldrew2024active} and XPO~\citep{xie2024exploratory}. We omit the comparison with SELM~\citep{zhang2024self} since SELM and XPO share a very similar algorithmic design.\looseness=-1

\textbf{Metrics}. We use the win rate of agent's responses against reference responses judged by the reward oracle as the performance metric. This metric can reflect both the agent's cumulative regret and anytime regret (i.e., average performance). In the E\&E setting, we measure the ``online'' win rate of the agent's dueling responses that are executed during experience collection and take the average. In the BAI setting, we measure the ``offline'' win rate by evaluating the latest agent's responses given a fixed set of holdout prompts periodically. We mainly focus on the BAI setting because crowdsourcing seems a major scenario for most practitioners, and present one set of experiments for comparing different exploration strategies in both settings. When the comparison is only made within a model scale, we report the relative win rate against the initial STF models. When the comparison is across scales (\cref{fig:3-scale-benchmark}~Left), we report the absolute win rate against the ground truth responses in the dataset.\looseness=-1

\textbf{Hyperparameters}. We set $\beta=0.1$ for DPO and $\beta=0.2$ for SLiC and find they are robust for all scales. We tune $\beta$ from $\{0.2, 0.3, 0.5, 1.0\}$ for IPO across scales and report the best performing results. We sample $M=20$ on-policy responses with a temperature $\eta=0.7$ during training, and use greedy decoding for offline evaluation (BAI's metric). We use the Adam optimizer with learning rate of $5e-7$ and cosine scheduling, and set the batch size to be $128$. We initialize the mixture ratio $\gamma$ of \sea~to be $1$ and adjust it to $0.7$ after a burn-in period of $1$k samples. To reproduce baselines (APL and XPO), we follow the recommended hyperparameters from their papers.\looseness=-1

\textbf{Statistical significance}. There are various factors to introduce randomness during online learning. We thus launch $3$ independent runs for every experiment with different random seeds. All the results are reported with mean and standard error to indicate their statistical significance.\looseness=-1

\textbf{Computational resources}. Experiments at all scales are conducted on a single machine with 8 A100 GPUs to run the learner and actors. We additionally host a separate remote server with workers spawned on 16 A100 GPUs for the oracle RM\footnote{We utilize the Kubernetes service for routing requests to multiple Mosec~\citep{yang2021mosec} instances.}, so that it can be queried by all concurrently running experiments. All experiments conducted for this research consume about 2 A100 GPU years.\looseness=-1

\section{Supplementary materials}
\label{sec:app:supp}
We include a comparison of prior works (\cref{tab.compare}) and an example of ChatGPT's active exploration (\cref{fig:gpt_exp}) in this section.\looseness=-1

\newcommand{\bluetext}[1]{\textcolor{cyan!60!black}{\transparent{0.9}#1}}

\newcommand{\pinktext}[1]{\highlight{#1}}
\newcommand{\purpletext}[1]{\textcolor{red!40}{\transparent{0.9}#1}}
\definecolor{table-blue}{rgb}{0.678, 0.847, 0.902}

\begin{table}[h]
    \centering
    {\scriptsize
    \begin{tabular}{ccccccccc}
        \toprule
         & \multirow{2}{*}{{Method}} & \multicolumn{2}{c}{Exploration} & \multicolumn{3}{c}{Interaction} & \multicolumn{2}{c}{Proposal Policy} \\
         \cmidrule(lr){3-4}\cmidrule(lr){5-7}\cmidrule(lr){8-9}
         & & Active & Passive & Online & Iterative & Offline & $\pi_\theta$ & $\pi_\beta$ \\
         \toprule
         \toprule
    \multirow{4}{*}{\parbox[c]{1.5cm}{\centering RL\\Optimizer}} 
         & \cite{christiano2017deep} &      &\cmark&  &\cmark &\cmark &\bluetext{\cmark}&  \\
         & \cite{stiennon2020learning} &      &\cmark&  &\cmark &\cmark &\bluetext{\cmark}&  \\
         & \cite{bai2022training} &      &\cmark&  &\cmark &\cmark &\bluetext{\cmark}&  \\
         & \cite{ouyang2022training} &      &\cmark&  &\cmark &\cmark &\bluetext{\cmark}&  \\
         \midrule
    \multirow{13}{*}{\parbox[c]{1.5cm}{\centering Direct\\Optimizer}}
         & \cite{zhao2023slic} &   &\cmark  &  &  &\cmark  &\bluetext{\cmark}  &         \\
         & \cite{rafailov2023dpo} &   &\cmark  &  &  &\cmark  &\bluetext{\cmark}  &         \\
         & \cite{azar2024ipo} &   &\cmark  &  &  &\cmark  &\bluetext{\cmark}  &         \\
         & \cite{meng2024simpo} &   &\cmark  &  &  &\cmark  &\bluetext{\cmark}  &         \\
         & \cite{xu2023iterativedpo} &   &\cmark  &  &\cmark  &  &\bluetext{\cmark}  &         \\
         & \cite{guo2024direct} &   &\cmark  &\purpletext{\cmark}  &  &  &\bluetext{\cmark}  &         \\
         & \cite{mehta2023sample} &\pinktext{\cmark}  &  &\purpletext{\cmark}  &  &  &  &\cmark         \\
         & \cite{das2024provably} &\pinktext{\cmark}  &  &\purpletext{\cmark}  &  &  &  &\cmark         \\
         & \cite{melo2024deep} &\pinktext{\cmark}  &  &\purpletext{\cmark}  &  &  &  &\cmark         \\
         & \cite{dwaracherla2024efficient} &\pinktext{\cmark}  &  &\purpletext{\cmark}  &  &  &  &\cmark         \\
         \rowcolor{table-blue!40} & \cite{zhang2024self} &\pinktext{\cmark}  &  &\purpletext{\cmark}  &  &  &\bluetext{\cmark}  &         \\
         \rowcolor{table-blue!40} &\cite{xie2024exploratory} &\pinktext{\cmark}  &  &\purpletext{\cmark}  &  &  &\bluetext{\cmark}  &         \\
         \rowcolor{table-blue!40} &  \cite{muldrew2024active} &\pinktext{\cmark}  &  &\purpletext{\cmark}  &  &  &\bluetext{\cmark}  &        
         \\
         \bottomrule
    \end{tabular}
    }
    \caption{A summary of prior work. $\pi_\theta$ denotes the proposal policy that is continuously updated based on newly collected preference data, while $\pi_\beta$ denotes a fixed proposal policy. Algorithms that encompass online interaction (\cref{property1}), active exploration (\cref{property2}), and learnable $\pi_\theta$ offer the best sample efficiency. Notably, only \colorbox{table-blue!40}{three methods} (listed at the bottom of the table) satisfy these characteristics, and we include them for comparisons in our experiments.\looseness=-1}
    \label{tab.compare}
\end{table}

\begin{figure}[b]
    \centering
    \includegraphics[width=\linewidth]{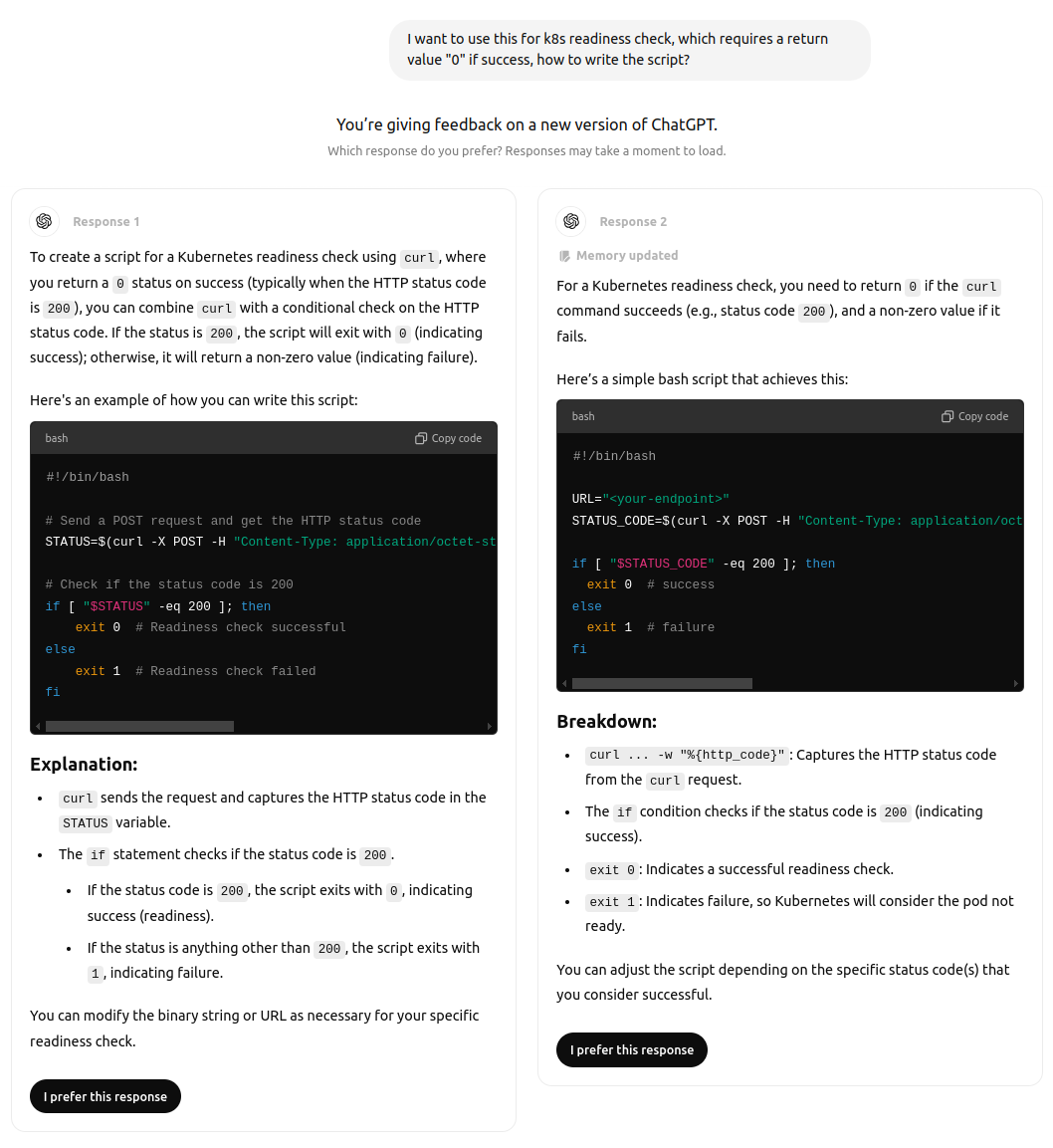}
    \caption{ChatGPT system asks for users' preference feedback to strategically explore better answers. In this case, algorithms should be designed around the objective of \textit{minimizing cumulative regret} (i.e., the E\&E setting), because the quality of both responses generated by the system affects user experience.\looseness=-1}
    \label{fig:chat-gpt-online-feedback}
\end{figure}

\end{document}